\documentclass[review]{elsarticle}

\usepackage{lineno,hyperref}
\usepackage[utf8]{inputenc}
\usepackage[T1]{fontenc}
\usepackage{algorithm} 
\usepackage{amssymb}
\usepackage{amsmath} 
\usepackage{amsthm}
\usepackage{dsfont}
\usepackage{hyperref}
\usepackage{graphicx}
\usepackage{subcaption}
\DeclareMathOperator*{\argmax}{arg\!\max}
\DeclareMathOperator*{\argmin}{arg\!\min}

\modulolinenumbers[5]

\usepackage{colortbl}
\def\R#1{\textcolor{black}{#1}}

\newcommand{\modif}[1]{{\color{black} {#1}}}

\newcommand{\cy}[1]{{\color{black}{#1}}}

\journal{Reliability Engineering \& System Safety }

\begin{document}

\begin{frontmatter}

\title{\modif{Importance sampling based active learning for parametric seismic fragility curve estimation}}

\author[1,2]{Clément Gauchy}
\author[1]{Cyril Feau}
\author[2]{Josselin Garnier}

\address[1]{Université Paris-Saclay, CEA, Service d'\'Etudes Mécaniques et Thermiques, 91191, Gif-sur-Yvette, France}
\address[2]{CMAP, \'Ecole Polytechnique, Institut Polytechnique de Paris, 91128 Palaiseau Cedex, France}

\begin{abstract}
\cy{The key elements of seismic probabilistic risk assessment studies are the fragility curves which express the probabilities of failure of structures conditional to a seismic intensity measure. A multitude of procedures is currently available to estimate these curves. For modeling-based approaches which may involve complex and expensive numerical models, the main challenge is to optimize the calls to the numerical codes to reduce the estimation costs. Adaptive techniques can be used for this purpose, but in doing so, taking into account the uncertainties of the estimates (via confidence intervals or ellipsoids related to the size of the samples used) is an arduous task because the samples are no longer independent and possibly not identically distributed. The main contribution of this work is to deal with this question in a mathematical and rigorous way. To this end, we propose and implement an active learning methodology based on adaptive importance sampling for parametric estimations of fragility curves. We prove some theoretical properties (consistency and asymptotic normality) for the estimator of interest. Moreover, we give a convergence criterion in order to use asymptotic confidence ellipsoids. Finally, the performances of the methodology are evaluated on analytical and industrial test cases of increasing complexity.}

\end{abstract}


\end{frontmatter}

\linenumbers

\section{Introduction}
\cy{The notion of fragility curve appeared in the 1980s as a key element of Seismic Probabilistic Risk Assessment (SPRA) studies (see e.g. \cite{Kennedy1980,KENNEDY198447, PARK1998}). A fragility curve expresses the probability of failure of a mechanical structure subjected to earthquake excitations conditional to a seismic Intensity Measure (IM), that may be the Peak Ground Acceleration (PGA) or the Pseudo-Spectral Acceleration for a given frequency (PSA). Fragility curves are also part of Performance-Based Earthquake Engineering (PBEE) framework \cite{GHOBARAH2001878, Porter2007} and are not limited to seismic loading. For example, they are also used for wind loading \cite{Quilligan12} or to address the problem of multi-hazard storm conditions \cite{BERNIER2019}.}

\cy{In earthquake engineering, various data sources can be exploited to estimate these curves, namely: expert judgments supported by test data \cite{Kennedy1980,KENNEDY198447,PARK1998,Zentner2017}, experimental data \cite{PARK1998, Gardoni2002}, post-earthquake damage results called empirical data \cite{STRAUB2008, Lallemant2015} and analytical results given by more or less refined numerical models (most of the works cited in this article fall into this category). Moreover, in practice, three families of procedures coexist to evaluate seismic fragility curves: Incremental Dynamic Analysis (IDA), Multiple Stripes Analysis (MSA) and Cloud Analysis (CA). In most cases, these approaches are also coupled with a parameterization of the fragility curve since, compared to non-parametric estimations, parametric ones require small sample sizes. The lognormal model historically introduced in the SPRA framework \cite{Kennedy1980,KENNEDY198447} is currently the most used (see e.g. \cite{Shinozuka00, ELLINGWOOD2001251, KIM2004, MANDAL201611, WangF2020}), even if its validity is questionable (see e.g. \cite{Mai2017,TREVLOPOULOS2019}).}

\cy{In a few words, IDA is based on scaled accelerograms until the failure threshold of interest. Fragility curves are then considered as empirical cumulative distribution functions. The main disadvantage of IDA is that excessive scaling can lead to signals with unrepresentative frequency content and duration, which can imply biased results in nonlinear structural responses \cite{Luco2007,ALTIERI2020}. Strong evidence against scaling accelerograms is provided in \cite{Grigoriu2011} considering a more theoretical approach. Although not recommended, this approach is still implemented (see e.g. \cite{MANDAL201611,ZHAO2021}). MSA is based on multiple accelerograms selected or scaled to match specific IMs. Thus, for each IM value, the structural analyses provide a fraction of the ground motions that cause failure. Finally, CA is a kind of generalization of MSA in the sense that it is based on a single accelerogram for each IM value.}

\cy{Depending on the context, different techniques can be employed to estimate the fragility curves. For example, for parametric estimation, we distinguish: the method of moments (which is mainly used for the IDA-based methodology), the Maximum Likelihood Estimation (MLE) by assuming the independence of the observations (which can be questionable when empirical data are concerned \citep{STRAUB2008}), and the minimization of the Sum of Squared Errors (SSE). For non-parametric estimation, kernel smoothing can be used (see e.g. \cite{Lallemant2015, Mai2017}) as well as other methodologies (see e.g. \cite{TREVLOPOULOS2019, ALTIERI2020}). Note that most of these strategies are compared in \cite{Lallemant2015, Mai2017, Baker2015} and \cite{Lallemant2015} gives a clear presentation of the advantages and disadvantages of each of them.}

\cy{Beyond these methods, techniques based on machine learning can also be used, including: linear regression or generalized linear regression \cite{Lallemant2015}, classification-based techniques \cite{BERNIER2019,KIANI2019,Sainct20}, kriging \cite{Gidaris15}, polynomial chaos expansion \cite{Mai16}, artificial neural networks \cite{Sainct20, MITROPOULOU2011, WANGZ2018}. Some of them were coupled with adaptive techniques to reduce the number of calculations to be performed \cite{Sainct20,Gidaris15}. The Bayesian framework is also relevant in this context since it allows either (i) to fit numerical models (metamodels, mathematical expressions based on engineering judgments, etc.) to experimental data to directly estimate the fragility curves~\cite{Gardoni2002} or (ii) to use empirical  data or analytical data to fit the parametric models of the fragility curves~\cite{STRAUB2008}.}

\cy{The major drawback of most of the work of the literature is that it does not address the issue of Confidence Intervals (CIs) related to the size of the samples used, in order to reflect the estimation uncertainty. This question arises, among others, when implementing computationally expensive numerical models of complex structures because very little data are then available. When data are assumed to be from an independent and identically distributed population, the bootstrap method can be used (see e.g. \cite{Shinozuka00, WangF2020, ZENTNER20101614}). However, when little data are involved, bootstrap samples can lead to unrealistic fragility curves such as unit-step functions. As a result, this can lead to excessively large CIs. Moreover, when active learning techniques are implemented, the samples are no longer independent and the bootstrap method cannot be used. It should be mentioned that several works have tackled the question of the convergence of the estimations according to the number of available data, but only from an empirical point of view since, in most cases, it seems that there is no other possibility (see e.g.~\cite{TREVLOPOULOS2019, ZHAO2021, Sainct20, ZENTNER20101614, Gehl2015}).}

\cy{For the modeling-based approaches, the aim of this work is to address the issue of optimized fragility curve estimations (i.e. based on few data) within a rigorous mathematical framework. To do this, we propose and implement an active learning methodology based on adaptive importance sampling \cite{Chu2011} in a statistical learning context \cite{hastie_09_elements-of.statistical-learning}, \modif{called Importance Sampling based Active Learning (IS-AL)}.}


\modif{Adaptive importance sampling was introduced for classical Monte Carlo integral approximation in \cite{Kloek1978}, and later studied in \cite{Oh1992}. \modif{Moreover, adaptive importance sampling is also used in industrial applications and} have already been discussed, implemented and tested for probability estimation  of rare event (e.g. failure state) in reliability analysis \cite{GONG2018199, Papaioannou2019}.} \cy{By applying it to the parametric estimations of the fragility curves, we show by asymptotic analysis and numerical simulations that IS-AL allows for (i) a rapid convergence of the estimated fragility curve towards the true (unknown) fragility curve and (ii) a rigorous quantification of the estimation uncertainty. It gives  asymptotic CIs and confidence ellipsoids for the quantities of interest as well as statistical tests to determine whether the asymptotic regime has been reached and whether asymptotic CIs and confidence ellipsoids can be used.}


\R{The proposed methodology relies on parametric approximations of fragility curves for any IM of interest. Although the validity of parametric models is both questionable and difficult to assess (see e.g. \cite{Mai2017, TREVLOPOULOS2019, ZENTNER201754}), some numerical experiments based on the seismic responses of simple mechanical systems - i.e few degrees of freedom systems - suggest that the choice of an appropriate IM makes it possible to reduce the potential biases between reference fragility curves - that can be obtained by massive Monte Carlo simulations - and their parametric approximations. This point is illustrated in the application section \ref{sec:NumRes} of this paper. Remember, however, that in practice, the selection of an optimal IM is not a trivial matter (see e.g. \cite{HARIRIARDEBILI201667, Ciano2020}) and Machine Learning techniques can be used for this purpose (e.g. \cite{Sainct20}), knowing that the references \cite{LucoCornell07} and \cite{Padgett08} give optimality criteria for selection of such IM. 
}

\cy{In this work, the methodology is applied to different test cases and compared with more traditional approaches such as MLE often used by practitioners (see e.g. \cite{STRAUB2008, Lallemant2015, Shinozuka00, WangF2020, Mai2017, Baker2015, ZENTNER20101614}). In order to avoid the scaling of the accelerograms, the stochastic model of modulated and filtered white-noise process defined in~\cite{Rezaeian10} is used to enrich a set of real ground motion records selected in a database using magnitude and distance
criteria. This stochastic model is chosen because it well encompasses the temporal and spectral non-stationarities of real seismic signals. Additionally, it has already been used in several works (see e.g. \cite{Mai2017, Sainct20, Kwong15}).}

\cy{The paper is organized as follows. In section \ref{section2} the statistical framework is defined for any parametric fragility curve model and any IM. Section \ref{section3} is dedicated to the presentation of the IS-AL algorithm applied to seismic fragility curves estimation for the lognormal model. Section \ref{sec:thres} summarizes the main theoretical results of this work, which are proved in the appendices. These results concern a criterion for evaluating the convergence of the IS-AL strategy and the definition of asymptotic confidence ellipsoids for the fragility curve parameters. Section \ref{sec:PerfEvalISAL} presents the performance metrics used in this work to compare IS-AL, random sampling and MLE strategies. Finally, in section \ref{sec:NumRes}, IS-AL performance is assessed on analytical and industrial test cases of increasing complexity.}

\section{Parametric seismic fragility curve estimation: a statistical learning framework}
\label{section2}%

\subsection{\cy{General framework}}

We consider the following situation. Let $\mathcal{X}$ be a compact set of $\mathbb{R}$, $X$ a $ \mathcal{X}$-valued random variable and $S \in \{0, 1\}$ a random label. \cy{In SPRA studies $X = \log \text{IM}$ -- more generally we can define $X = \psi(\text{IM})$ where $\psi$ is an increasing function of IM such as a Box-Cox transform \cite{boxcox} -- }
and $S$ is the indicator variable of the failure of the structure. The pair $(X,S)$ has the probability distribution $P$ over $\mathcal{X} \times \{0,1\}$:
\begin{equation}
P(dx, ds) = \big[ \mu(x) \delta_{1}(ds) + (1 - \mu(x)) \delta_{0}(ds) \big] p(x) dx \ ,
\end{equation}
where $\delta_j$ is the Dirac distribution at $j$, 
$p$ is the marginal probability density function (pdf) of $X$,
and the fragility curve $\mu(x)$ is the conditional expectation of $S$ (conditional probability of failure or fragility curve): \begin{equation}
\mu(x) = \mathbb{E}[S|X = x] \ .
\end{equation}
The aim of the paper is to estimate the curve $\mu(x)$ from datapoints $(X_i, S_i)_{i=1}^n$ that may be independent and identically distributed with the distribution $P$ or that may be selected by a more appropriate scheme. 
\cy{As mentioned in the introduction}, it is a classical assumption to use a parametric form for the fragility curve $\mu$ to tackle the need for time consuming mechanical simulations, we thus consider the space of functions $\mathcal{F} = \{f_{\theta}, \theta \in \Theta \}$, where $x\mapsto f_\theta(x)$ is a function from $\mathbb{R}$ to $[0,1]$ for any $\theta$ and $\Theta\subset \mathbb{R}^m$. The goal is to minimize the quadratic risk:
\begin{equation}
g(\theta) = \mathbb{E}[(\mu(X) - f_{\theta}(X))^2] \: ,
\end{equation}
in order to find (provided it exists and is unique):
\begin{equation}
    \label{def:thetastar}
\theta_{*} = \argmin_{\theta \in \Theta} g(\theta) \ .
\end{equation}
Unfortunately, the observable data are $(X_i, S_i)_{i=1}^n$, we do not observe directly $\mu(X_i)$. 
But considering that:
\modif{
\begin{align}
\nonumber
        \mathbb{E}[(S - f_{\theta}(X))^2] &= \mathbb{E}[(S - \mu(X))^2] +  \mathbb{E}[(\mu(X) - f_{\theta}(X))^2]\\
\nonumber
        & \quad + 2\mathbb{E}[(\mu(X) - f_{\theta}(X))(S - \mu(X))] \\
         &=  \mathbb{E}[\mu(X)(1-\mu(X))] + \mathbb{E}[(\mu(X) - f_{\theta}(X))^2]
\label{eq:expanderror}
\end{align}
because $\mathbb{E}[S^2|X]=\mathbb{E}[S|X]=\mu(X)$,
}
we can observe that the minimization with respect to $\theta$ of $\mathbb{E}[(S - f_{\theta}(X))^2]$ is equivalent to the minimization  of $\mathbb{E}[(\mu(X) - f_{\theta}(X))^2]$. 
Hence, we will consider the quadratic risk 
\begin{equation}
r(\theta) = \mathbb{E}[(S - f_{\theta}(X))^2] \ .
\end{equation}

In the context of classical learning, when we observe $n$ datapoints $(X_i, S_i)_{i=1}^n$ drawn independently from the probability distribution $P(dx,ds)$ over $\mathcal{X} \times \{0,1\}$,
the expectation can be approximated by the empirical mean:
\begin{equation}
\label{eq:defsquaresloss}
\widehat{R}_n(\theta) = \frac{1}{n} \sum\limits_{i=1}^n (S_i - f_{\theta}(X_i))^2 \: .
\end{equation}
The corresponding passive estimator (the term passive is used to highlight the absence of any particular sampling strategy) is then:
\begin{equation}
\widehat{\theta}_n  = \argmin_{\theta \in \Theta} \widehat{R}_n(\theta).
\label{R_RS}
\end{equation}

Conversely to classical learning, active learning aims at selecting the most useful numerical experiments to be carried out in order to form the learning set. In the passive strategy, the datapoints $X_i$ are sampled from the original probability distribution with pdf $p$ \modif{drawn from 
a stochastic ground-motion model}.
\modif{In the same way as in \cite{Chu2011}, we propose an} active learning strategy, called \modif{Importance Sampling based Active Learning (IS-AL)}. It consists to draw the datapoints $X_i$ from an instrumental probability distribution with \R{pdf $q$} that is chosen in an adaptive way. 
\R{In our context, it is} straightforward to use a rejection method \modif{applied to the stochastic ground-motion model} in order to generate seismic loads with a desired intensity measure distribution. \cy{Let us recall in fact that the label $S_i$ (which gives the failure state of the structure) is, in our case, expensive to obtain because it comes from complex numerical simulations of mechanical structures while the artificial seismic signals are inexpensive to generate.}

The main \modif{objective} of this procedure is to reduce the variance implied by the empirical approximation of the quadratic risk $r(\theta)$. Importance Sampling is a classical variance reduction technique for Monte Carlo estimation used in structural reliability \cite{Papaioannou2019,Zuniga2021}.  
\cy{If the $(X_i)_{i=1}^n$ are sampled with the pdf $q$ and $(S_i)_{i=1}^n$ are the labels obtained from $n$ calls to the mechanical model,}
\cy{then the importance sampling estimator of the empirical quadratic risk is:}
\begin{equation}
\label{eq:hatRnq}
\widehat{R}^{\text{IS}}_{n}(\theta) = \frac{1}{n} \sum\limits_{i=1}^n \frac{p(X_i)}{q(X_i)} (S_i - f_{\theta}(X_i))^2.
\end{equation}
\R{In the rest of the paper,} we will denote 
by $r(\theta) = \mathbb{E}_{(X, S) \sim P}[\ell_{\theta}(X, S)]$ with $\theta \mapsto \ell_{\theta}(x, s)$ a positive loss function for the sake of generalization. For the numerical applications, only the case of the quadratic loss 
\begin{equation} \label{eq:quadloss}
l_{\theta}(x, s) = (s - f_{\theta}(x))^2,
\end{equation}
will be considered.

\subsection{\cy{Problem regularization for the parametric lognormal model}}

\modif{For applications to seismic fragility curves estimation, a classical space of functions to approximate $\mu$ is $\mathcal{F} = \{ \Phi( \frac{ \log( {\rm IM} /\alpha) }{ \beta} ), (\alpha, \beta) \in \Theta \}$ where $\Phi$ is the cumulative distribution function of the standard Gaussian distribution \cite{Kennedy1980}, $\theta = (\alpha,\ \beta)^T$, and $\Theta$ a compact set of $(0,+\infty)^{2}$ (therefore $m=2$). 
Compactness of $\Theta$ is a common assumption in our applications. From an engineer perspective, it is possible to bound $\alpha$ and $\beta$.
\cy{However, in practice, the lower bound for $\beta$ may be reached by the different estimators. Consequently, inspired by Bayesian inference theory \cite{Box1973}, we introduce a regularization term
$\Omega(\theta;\beta_{\rm reg})$ to tackle this issue (we will take $\Omega(\theta;\beta_{\rm reg}) = \beta_{\rm reg}/ \beta$ below). The squared loss (\ref{eq:defsquaresloss}) is then replaced by:}
\begin{equation}
    \widehat{R}_{n,{\rm reg}}(\theta;\beta_{\rm reg}) = \frac{1}{n} \sum\limits_{i=1}^n \cy{\ell_{\theta}(X_i, S_i)} + \frac{\Omega(\theta;\beta_{\rm reg})}{n} \ .
\end{equation}
The derivation of the importance sampling estimator of the regularized square loss is straightforward:
\begin{equation}
    \widehat{R}^{\rm IS}_{n, {\rm reg}}(\theta;\beta_{\rm reg}) = \frac{1}{n} \sum\limits_{i=1}^n \frac{p(X_i)}{q(X_i)} \cy{\ell_{\theta}(X_i, S_i)} + \frac{\Omega(\theta;\beta_{\rm reg})}{n} \ .
    \label{eq:hatRnqreg}
\end{equation}
This regularization is motivated by the intrinsic difficulty of estimating the \cy{standard deviation} $\beta$ of the lognormal model when $\beta$ is small \cite{Keller15}. Fragility curves with small $\beta$ are hard to distinguish due to the convergence towards a degenerate $0-1$ fragility curve.}

\section{\cy{Principles of the IS-AL strategy}}
\label{section3}

\cy{This section focuses on the choice of an optimal density $q$ (section~\ref{sec:choiceq}) as well as on the description of the IS-AL strategy (section~\ref{sec:ISALprinciple}).}

\subsection{\cy{Choice of an optimal density $q$} \label{sec:choiceq}}

\R{The heuristic} used to find a good instrumental probability distribution family is presented in \cite{Chu2011}. The first idea would be to minimize the variance of the importance sampling risk estimator (\ref{eq:hatRnqreg}):
\begin{align}
 {\rm Var}\big(\widehat{R}^{\rm IS}_{n, {\rm reg}}(\theta) \big) = \frac{1}{n} \Big\{ \iint_{\mathcal{X} \times \{0,1\}} \frac{p(x)}{q(x)} \ell_{\theta}(x, s)^2 P(dx,ds) - r(\theta)^2\Big\} \ ,
\end{align}
with respect to $q$ within the set of all pdfs. 
If we denote by $\Tilde{\ell}^2_{\theta}(x) =  \mathbb{E}[\ell_{\theta}(X, S)^2 | X = x]$ the \modif{squared} loss averaged on $S$:
\begin{equation}
\label{eq:tildeltheta}
    \Tilde{\ell}^2_{\theta}(x) = \mu(x)\ell_{\theta}(x, 1)^2 + (1 - \mu(x)) \ell_{\theta}(x, 0)^2 \ ,
\end{equation}
the variance of the importance sampling risk estimator (\ref{eq:hatRnqreg}) can be expressed as
$$
 {\rm Var}\big(\widehat{R}^{\rm IS}_{n, {\rm reg}}(\theta) \big) = \frac{1}{n}\Big\{ \int_{\mathcal{X}} \frac{p(x)^2}{q(x)} \Tilde{\ell}^2_{\theta}(x)  dx - r(\theta)^2 \Big\}\ ,
$$
and we look for
\begin{equation}
q_{\theta}^* = \argmin_q \int_{\mathcal{X}} \frac{p(x)^2}{q(x)}\Tilde{\ell}^2_{\theta}(x)dx  \ .
\end{equation}
Using \modif{Jensen's inequality \cite[Theorem 3.12]{Robert2005}}, we can solve the optimization problem and we can find that the optimal sampling pdf is of the form
$$
q_{\theta}^*(x) \propto \Tilde{\ell}_{\theta}(x)p(x) \ ,
$$
which depends on $\mu$ because $\Tilde{\ell}_{\theta}$ depends on $\mu$
[Here and below $\propto$ means equality up to a multiplicative constant]. Hence an approximation step is made by replacing $\mu$ by $f_{\theta}$ in (\ref{eq:tildeltheta}):
\begin{equation}
    \Tilde{\ell}^2_{\theta}(x) \approx f_{\theta}(x)\ell_{\theta}(x, 1)^2 + (1 - f_{\theta}(x)) \ell_{\theta}(x, 0)^2 \ .
\end{equation}
Hence the instrumental density becomes:
\begin{equation}
    q_{\theta}(x) \propto p(x)\sqrt{f_{\theta}(x)\ell_{\theta}(x, 1)^2 + (1 - f_{\theta}(x)) \ell_{\theta}(x, 0)^2} \ .
\end{equation}
Note that the instrumental distribution depends \R{on} $\theta$, the parameter we aim to estimate. \modif{Moreover, using IS-AL with the instrumental density $q_{\theta}$ directly could increase the variance if the density has light tails. We propose finally a defensive strategy as illustrated in \cite{Owen00, Hesterberg95}. The instrumental density becomes
\begin{equation}
q_{\theta, \varepsilon}(x) = \varepsilon p(x) + (1 - \varepsilon) q_{\theta}(x) ,
\label{eq:qteps}
\end{equation}
with $\varepsilon \in [0, 1]$. $\varepsilon$ is a mixing parameter, between the original marginal pdf $p(x)$ and the instrumental one $q_{\theta}(x)$, meaning that one time out of ${1}/{\varepsilon}$ the element is drawn from the pdf $p(x)$. This distribution allows to bound the likelihood ratio:
\begin{equation}
    \frac{p(x)}{q_{\theta, \varepsilon}(x)} = \frac{1}{\varepsilon + (1 - \varepsilon)\frac{q_{\theta}(x)}{p(x)}} < \frac{1}{\varepsilon} \ .
\end{equation}
Thus the defensive strategy bounds the variance even if the likelihood ratio ${p(x)}/{q_{\theta}(x)}$ is large.} 

\subsection{\cy{Description of the IS-AL strategy} \label{sec:ISALprinciple}}

\subsubsection{\cy{Algorithm}}

The procedure for computing the IS-AL estimator $\modif{\widehat{\theta}^{\rm IA}_n}$ is described in Algorithm \ref{alg:IS-AL}. \cy{Its main objective is to use an updated instrumental density $q_{\theta,\varepsilon}$ at each step. Note that (i) the algorithm needs to start from a certain parameter value $\modif{\widehat{\theta}_{0}^{\rm IA}}$ and (ii) we choose $\Omega (\theta; \beta_{\rm reg}) =  {\beta_{\rm reg}}/{\beta}$ for the regularization term in equation (\ref{eq:hatRnqreg}).} 

\begin{algorithm}[!ht]
    \caption{\modif{\cy{Importance Sampling based Active Learning (IS-AL)}}}\label{alg:IS-AL}
    \begin{enumerate}
    \item \cy{Choice of $\widehat{\theta}_{0}^{\rm IA}$ (section~\ref{sec:ISALinit}) and estimations of $\beta_{\rm reg}$ and $\varepsilon$ (section~\ref{sec:ISALbeta}).}
    \item For $i=1,\ldots,n$:
        \begin{enumerate}
        \item Draw $X_i$ from the distribution with pdf $q_{\modif{\widehat{\theta}_{i-1}^{\rm IA}}, \modif{\varepsilon}}$.
        \item Call the mechanical simulation at point $X_i$ to get label $S_i$
        \item Compute
            \begin{align}\label{eq: is-al}
            \modif{\widehat{\theta}_{i}^{\rm IA}} & = \argmin_{\theta \in \Theta} \widehat{R}^{\rm IA}_{i,{\rm reg}}(\theta;\beta_{\rm reg}), \\
            \widehat{R}^{\rm IA}_{i,{\rm reg}}(\theta;\beta_{\rm reg}) &  = 
            \frac{1}{i} \sum\limits_{j=1}^i \frac{p(X_j)}{q_{\modif{\widehat{\theta}_{j-1}^{\rm IA}}, \modif{\varepsilon}}(X_j)} \ell_{\theta}(X_j, S_j) + \frac{\beta_{\rm reg}}{n\beta} \label{eq:RIA}.
            \end{align}
        \end{enumerate}
    \end{enumerate}
    \end{algorithm}
    \cy{Additionally, a convergence criterion is presented in section~\ref{sec:convergence crit} and an asymptotic confidence ellipsoid for $\theta_*$ centered on $\widehat{\theta}_{n}^{\rm IA}$ is defined by equation~(\ref{eq:asymp ce}).}



\subsubsection{\cy{Initialization and choice of $\widehat{\theta}^{\rm IA}_0$} \label{sec:ISALinit}}

\cy{Regarding the initialization, as expected, the closer $\widehat{\theta}^{\rm IA}_0$ is from the true parameter $\theta_*$ the faster IS-AL is in asymptotic normal regime. A naive approach is to get a small sample of size $n_0$ (e.g. $n_0=20$) $(X_i, S_i)_{i=1}^{n_0}$ from the original marginal density $p$ of $X$ and to compute the passive learning estimator $\widehat{\theta}_{n_0}$ (equation~(\ref{R_RS})). This crude estimation can be used as the initial parameter ${\widehat{\theta}^{\rm IA}_0}$ to start IS-AL.

A better approach is to consider a metamodel - in the broad sense - of the mechanical simulation. As often used by practitioners, a numerical resolution based on a modal base projection can be implemented to get an estimate of the fragility curve corresponding to the linear behavior of the structure of interest. It is then possible to get a huge amount of datapoints  of the reduced model  (e.g. an independent and identically distributed sample of $n_{\rm red}=10^3$--$10^5$ pairs $(X_i,S_{{\rm red},i})_{i=1}^{n_{\rm red}}$ where $X_i$ is sampled with the original pdf $p$ and $S_{{\rm red},i}$ is the associated label obtained with the reduced model). The initial parameter ${\widehat{\theta}_{0}^{\rm IA}}$ is then chosen to be equal to $\widehat{\theta}_{n_{\rm red}}$. Statistical metamodels could also be used such as Gaussian Processes \cite{ECHARD2013232} or Support Vector Machines \cite{Sainct20}. 

In our applications reduced models are only used to give us prior knowledge on the fragility curve shape, encapsulated in the initial parameter of the {IS-AL} procedure. We then initialize {IS-AL} with a small sample of $20$ datapoints with the instrumental density 
${q}_{\modif{\widehat{\theta}}^{\rm IA}_0,\varepsilon}$ (equation~(\ref{eq:qteps})).
In other words, in Step 2 of Algorithm \ref{alg:IS-AL}, we do not update $\widehat{\theta}^{\rm IA}_i$ during the first $20$ steps.}


\subsubsection{\cy{Estimations of $\beta_{\rm reg}$ and $\varepsilon$ \label{sec:ISALbeta}}}

\cy{The regularization parameter, called $\widehat{\beta}^{\rm \cy{IA}}_{\rm reg}$, is determined by minimizing the Leave One Out error on the initialization sample (see previous section).}

\cy{Regarding the choice of the defensive parameter value $\varepsilon$, it is cumbersome and there is no direct methodology for its estimation. Moreover, its value depends strongly of the problem studied as shown in \cite{Bect2015}. Nevertheless, in section ~\ref{sec:osciNL}, we propose a benchmark in order to evaluate the "optimal" value of $\varepsilon$ for the class of structures considered in this study.}

\section{\cy{Theoretical results}}
\label{sec:thres}

\cy{This section summarizes the main theoretical results of this work. Section~\ref{sec:CandAC} addresses the issue of the consistency and asymptotic normality for the IS-AL estimator. Then, in section~\ref{sec:convergence crit}, a convergence criterion is proposed in order to be able to use the asymptotic confidence ellipsoids defined in section~\ref{sec:ACE}. A discussion is finally proposed about the practical use of the convergence criterion in section~\ref{sec:discussion}.}

\subsection{\cy{Consistency and asymptotic convergence of the IS-AL estimator} \label{sec:CandAC}}
We derive some theoretical properties for the estimator $\modif{\widehat{\theta}_{n}^{\rm IA}}$, consisting in its consistency \modif{towards the parameter $\theta_*$ defined by (\ref{def:thetastar})} and its asymptotic normality \modif{by adapting several proofs of \cite{Deylon18} about asymptotic optimality of adaptive importance sampling}. Detailed proofs of the following \modif{results} are given in the Appendix. \modif{The proofs are given in a more general context of empirical risk minimization, instead of IS-AL specifically. Indeed, we consider that these theoretical results can be used in a broader manner for other kind of applications.}

\modif{We first prove in \ref{app: consistency} the consistency of the IS-AL estimator $\modif{\widehat{\theta}_n^{\rm IA}}$ using Algorithm~\ref{alg:IS-AL}:
\begin{equation}\label{eq: consistency}
    \modif{\widehat{\theta}}_n^{\rm IA} \xrightarrow[n \rightarrow +\infty]{} \theta_{*} \: \text{in probability} \ .
\end{equation}

Then, we prove in \ref{app: normality} the convergence of $\sqrt{n}(\modif{\widehat{\theta}}^{\rm IA}_n  - \theta_*)$ to a Gaussian random variable with mean zero and covariance matrix:
\begin{equation}\label{eq: covariance}
    G_{\theta_*, \varepsilon} = \ddot{r}(\theta_{*})^{-1}V(q_{\theta_{*}, \varepsilon})(\ddot{r}(\theta_{*})^{-1})^{T} \, ,
\end{equation}
where
\begin{equation}
   V(q_{\theta_{*},\varepsilon}) = \mathbb{E}\left[\frac{p(X)}{q_{\theta_{*},\varepsilon}(X)}
   \ell_{\theta_{*}} (X,S) \nabla f_{\theta_{*}}(X)\nabla f_{\theta_{*}}(X)^T \right] \ ,
\end{equation}
and $\ddot{r}(\theta_{*})$ is the Hessian of $r(\theta)$ at $\theta_*$.
}

A straightforward corollary of equation~(\ref{eq: covariance}) is that, if $G_{\theta_*,\varepsilon}$ is nonsingular (which we assume from now on), then for any $\xi \in (0,1)$: 
\begin{equation}
    \mathbb{P}\left(n(\modif{\widehat{\theta}}_n^{\rm IA} - \theta_*)^T G_{\theta_*,\varepsilon}^{-1}(\modif{\widehat{\theta}}_n^{\rm IA} - \theta_*) < q^{\xi}_{\chi^2(m)}\right) \xrightarrow[n \rightarrow +\infty]{} \xi \, ,
\end{equation}
with 
$q^{\xi}_{\chi^2(m)}$ the $\xi$-quantile of the $\chi^2(m)$ distribution (remember that $\theta=(\alpha,\beta)^T$ and $m=2$ for the lognormal model). Remark that the matrix $G_{\theta_*,\varepsilon}$ depends on the unknown parameter $\theta_*$. It is thus possible to use a plug-in estimator:
\begin{equation}\label{eq:Gn}
\widehat{G}_n = \widehat{\ddot{r}}_n(\modif{\widehat{\theta}_n^{\rm IA}})^{-1} \widehat{V}_n ( \modif{\widehat{\theta}_n^{\rm IA}} ) (\widehat{\ddot{r}}_n(\modif{\widehat{\theta}_n^{\rm IA}})^{-1})^T \ ,
\end{equation}
with 
\begin{equation}\label{eq: ddot_r_hat}
    \widehat{\ddot{r}}_n(\theta) = \frac{1}{n}\sum\limits_{i=1}^n \frac{p(X_i)}{q_{\modif{\widehat{\theta}}_{i-1}^{\rm IA}, \varepsilon}(X_i)} \ddot {\ell}_{\theta}(X_i, S_i) \ ,
\end{equation}
\begin{equation}\label{eq: V_hat}
\widehat{V}_n( \theta ) = \frac{1}{n} \sum\limits_{i=1}^n  \frac{p(X_i)^2}{q_{\theta,\varepsilon}(X_i)q_{\modif{\widehat{\theta}}_{i-1}^{\rm IA}, \varepsilon}(X_i)} \dot\ell_{\theta}(X_i, S_i)\dot\ell_{\theta}(X_i, S_i)^T \, ,
\end{equation}
and $\ddot\ell_{\theta}(x, s)$ the Hessian of $\ell_{\theta}(x,s)$ with respect to $\theta$. \modif{We have:
\begin{equation}\label{eq: Gn consistency}
\widehat{G}_n^{-1} \rightarrow G_{\theta_*,\varepsilon}^{-1} \ \text{in probability} .
\end{equation}}
The proof is in \ref{app:prooflemmaGn}. \modif{Using asymptotic normality of $\modif{\widehat{\theta}}_n^{\rm IA}$, we can show that}: $n(\modif{\widehat{\theta}}_n^{\rm IA} - \theta_*)^T G_{\theta_*,\varepsilon}^{-1}(\modif{\widehat{\theta}}_n^{\rm IA} - \theta_*) \rightarrow \chi^2(m)$. Using Slutsky's lemma, we have the following convergence in distribution:
\begin{equation}\label{eq: slutsky}
    n(\modif{\widehat{\theta}}_n^{\rm IA} - \theta_*)^T\widehat{G}_n^{-1}(\modif{\widehat{\theta}}^{\rm IA}_n - \theta_*) \xrightarrow[n \rightarrow +\infty]{}\chi^2(m) \ .
\end{equation}


\subsection{\cy{Convergence criterion using a statistical hypothesis test}\label{sec:convergence crit}}

The estimation of the generalization error without a validation set is often based on Cross Validation.
When IS-AL is used, the data points $(X_i, S_i)$ are no longer independent and identically distributed. 
We propose to use a \cy{convergence} criterion that ensures that asymptotic normality is reached. \modif{Consider two independent datasets $\mathcal{D}_1 = (X_{i,1}, S_{i,1})_{i=1}^{n}$ and $\mathcal{D}_2 = (X_{i,2}, S_{i,2})_{i=1}^{n}$ generated with IS-AL. Let $\widehat{R}^{\rm IA}_{n,{\rm reg}, j}$ be the weighted loss for $\mathcal{D}_j$ for $j=1,2$ defined as in (\ref{eq: is-al}). Denote:
$$
\modif{\widehat{\theta}}_{n,j}^{\rm IA} = \argmin_{\theta \in \Theta} \widehat{R}^{\rm IA}_{n,{\rm reg}, j}(\theta; \beta_{\rm reg}), \quad \quad j=1,2\,.
$$
Then we have:
\begin{equation}\label{eq: asymp grad}
\sqrt{n}(\dot{\widehat{R}}^{\rm IA}_{n,{\rm reg}, 1}(\modif{\widehat{\theta}}_{n,2}^{\rm IA};\beta_{\rm reg}) - \dot{\widehat{R}}^{\rm IA}_{n,{\rm reg}, 2}(\modif{\widehat{\theta}}_{n,1}^{\rm IA}; \beta_{\rm reg})) \xrightarrow{\mathcal{L}} \mathcal{N}(0, 8V(q_{\theta_{*},\varepsilon}, \dot{\ell}_{\theta_{*}}))
\end{equation}
as $n\to +\infty$.}
Denote
\begin{align}
\widehat{W}_n &= \frac{n}{8} (\dot{\widehat{R}}^{\rm IA}_{n,{\rm reg}, 1}(\modif{\widehat{\theta}_{n,2}^{\rm IA}}; \beta_{\rm reg}) - \dot{\widehat{R}}^{\rm IA}_{n,{\rm reg}, 2}(\modif{\widehat{\theta}_{n,1}^{\rm IA}}; \beta_{\rm reg}))^T \widehat{V}_{n,12}^{-1} (\dot{\widehat{R}}^{\rm IA}_{n,{\rm reg}, 1}(\modif{\widehat{\theta}_{n,2}^{\rm IA}}) - \dot{\widehat{R}}^{\rm IA}_{n,{\rm reg}, 2}(\modif{\widehat{\theta}_{n,1}^{\rm IA}}))
,\\
\widehat{V}_{n,12}&= \frac{1}{2}\big( \widehat{V}_{n, 1}(\modif{\widehat{\theta}_{n, 1}^{\rm IA}})+ \widehat{V}_{n, 2}(\modif{\widehat{\theta}_{n, 2}^{\rm IA}}) \big),
\label{eq: V_hat_12}
\end{align}
with $\widehat{V}_{n, j}$ the empirical estimator in equation \eqref{eq: V_hat} for the $j$-th IS-AL dataset $\mathcal{D}_j$ for $j=1,2$.

By equation \eqref{eq: asymp grad} and by Slutsky's lemma, $\widehat{W}_n$ converges weakly to $\chi^2(m)$. It is, therefore,  possible to define a convergence criterion inspired by statistical test theory to check the asymptotic normality of $\modif{\widehat{\theta}}^{\rm IA}_n$.
Our convergence criterion is equivalent to the hypothesis test:
\begin{equation}
    \begin{array}{ccc}
         (\mathcal{H}_0) \ : \widehat{W}_n \ \text{follows} \ \chi^2(m)    & \text{against} & (\mathcal{H}_1) \ : \widehat{W}_n \ \text{does not follow} \ \chi^2(m) \ . \\
    \end{array}
\end{equation}
For $\xi \in (0, 1)$, we then consider the statistical test which rejects $(\mathcal{H}_0)$ if:
\begin{equation}
    \widehat{W}_n > q^{\chi^2(m)}_{1 - \xi} \ ,
\end{equation}
where $q^{\chi^2(m)}_{1 - \xi}$ denotes the $(1 - \xi)$-quantile of the $\chi^2(m)$ distribution. Hence, this statistical test is of asymptotic level $\xi$. 

\subsection{\cy{Asymptotic confidence ellipsoid}\label{sec:ACE}}

\cy{Thanks to the equation \eqref{eq: slutsky}, it is possible to construct an asymptotic confidence ellipsoid of level $\xi  \in (0, 1)$ defined by:
\begin{equation}\label{eq:asymp ce}
    \mathcal{E}_{n, \xi}^{\rm IA} = \{\theta : n(\theta - \widehat{\theta}_n^{\rm IA})^T \widehat{G}_n^{-1}(\theta - \widehat{\theta}_n^{\rm IA}) < q^{\chi^2(m)}_{1 - \xi}\} \ ,
\end{equation}
with:
$$
\mathbb{P}(\theta_* \in \mathcal{E}_{n, \xi}^{\rm IA}) \xrightarrow[n \rightarrow +\infty]{} \xi.
$$

Because the convergence criterion $\widehat{W}_n$ indicates when the estimator follows the asymptotic Gaussian distribution, it also indicates at which sample size $n$ the value $\mathbb{P}(\theta_* \in \mathcal{E}_{n, \xi}^{\rm IA})$ is close to its theoretical value $\xi$.} 

\subsection{\cy{Discussion about the practical use of the convergence criterion} \label{sec:discussion}}

An apparent drawback of this convergence criterion is that it doubles the computational cost, due to the necessity of having two independent IS-AL estimators $\modif{\widehat{\theta}_{n,1}^{\rm IA}}$ and $\modif{\widehat{\theta}_{n,2}^{\rm IA}}$ to compute $\widehat{W}_n$. 
It is, however, possible to use the estimator 
\begin{equation}
    \modif{\widehat{\theta}_{n, 12}^{\rm IA}} = \frac{\modif{\widehat{\theta}_{n,1}^{\rm IA}} + \modif{\widehat{\theta}_{n,2}^{\rm IA}}}{2},
\end{equation}
which has an asymptotic variance that is half the one of $\modif{\widehat{\theta}_{n,1}^{\rm IA}}$ and $\modif{\widehat{\theta}_{n,2}^{\rm IA}}$.
\modif{Indeed, it is straightforward that $\sqrt{n}(\modif{\widehat{\theta}_{n, 12}^{\rm IA}} - \theta_*)$ converges in distribution to a zero mean Gaussian random variable with covariance matrix $G_{\theta_*, \varepsilon}/2$. 
It is, therefore, possible to define an asymptotic confidence ellipsoid 
which exploits all the data points used to build the estimator $\widehat{\theta}_{n, 12}^{\rm IA}$ of $\theta_*$:
$$
\mathcal{E}_{n,12, \xi}^{\rm IA} = \{\theta : 2n(\theta - \widehat{\theta}_{n,12}^{\rm IA})^T \widehat{G}_{n,12}^{-1}(\theta - \widehat{\theta}_{n,12}^{\rm IA}) < q^{\chi^2(m)}_{1 - \xi}\} ,
$$
with $\widehat{G}_{n,12}=\widehat{\ddot{r}}_{n,12}^{-1} \widehat{V}_{n,12} 
(\widehat{\ddot{r}}_{n,12}^{-1})^T $,
$\widehat{V}_{n,12}$ defined by (\ref{eq: V_hat_12}), 
$\widehat{\ddot{r}}_{n,12}=\frac{1}{2} \widehat{\ddot{r}}_{n,1} (\widehat{\theta}_{n,1}^{\rm IA})+\frac{1}{2} \widehat{\ddot{r}}_{n,2}(\widehat{\theta}_{n,2}^{\rm IA})$
and $\widehat{\ddot{r}}_{n,j}$ defined as (\ref{eq: ddot_r_hat}) with the dataset ${\cal D}_j$, $j=1,2$.
}

\section{\cy{Performance evaluation of the IS-AL strategy compared to the random sampling and MLE strategies}}
\label{sec:PerfEvalISAL}%

\cy{This section explains how to assess the performance of IS-AL with respect to Random Sampling (RS) and MLE strategies. In section~\ref{sec:RS_MLE}, RS and MLE principles are briefly summarized. Performance metrics inspired from \cite{Chabridon2017, Morio2015} to check the quality of IS-AL strategy are detailed in section~\ref{sec:PerfM}. Finally, the statistical procedure used to assess the quality of the IS-AL asymptotic confidence ellipsoid compared to that of a classical approach such as MLE is given in section~\ref{sec:ACEbench}.}

\subsection{\cy{RS and MLE principles \label{sec:RS_MLE}}}

\cy{RS strategy consists in applying the IS-AL algorithm with the proposal probability density $q$ being the marginal probability density $p$ of the intensity measure. This boils down to classical empirical risk minimization for supervised learning. The RS estimator $\widehat{\theta}_{n}^{\rm RS}$ is then defined by:}
\cy{\begin{align}\label{eq: is-al2}
\widehat{\theta}_{n}^{\rm RS} & = \argmin_{\theta \in \Theta} \widehat{R}^{\rm RS}_{n,{\rm reg}}(\theta;\beta_{\rm reg}), \\
\widehat{R}^{\rm RS}_{n,{\rm reg}}(\theta;\beta_{\rm reg}) &  = 
\frac{1}{n} \sum\limits_{i=1}^n \ell_{\theta}(X_i, S_i) + \frac{\beta_{\rm reg}}{n\beta} \label{eq:RS}.
\end{align}}

\cy{As mentioned in the introduction, MLE is a classical estimation method in the field of seismic probabilistic risk assessment and fragility curve estimation (see e.g. \cite{STRAUB2008, Lallemant2015, Shinozuka00, WangF2020, Mai2017, Baker2015, ZENTNER20101614}). It is defined by the estimator $\widehat{\theta}_n^{\rm MLE}$ that maximizes the likelihood given a dataset \cy{$(X_i, S_i)_{i=1}^{n}$} that is sampled at random from the original marginal density $p$ of $X$:
\begin{equation}
\widehat{\theta}_n^{\rm MLE} = \argmax_{\theta \in \Theta} \sum\limits_{i=1}^n S_i\log(f_{\theta}(X_i)) + (1 - S_i)\log(1 - f_{\theta}(X_i))     .
\end{equation} 

The initializations of the RS and MLE algorithms are based on 20 data points drawn at random from the original distribution $p$. For the RS algorithm, the regularization parameter, called $\widehat{\beta}^{\rm RS}_{\rm reg}$, is computed using Leave One Out cross validation as for the IS-AL algorithm.}


\subsection{Performance metrics \cy{for} the numerical benchmarks \label{sec:PerfM}}
\cy{This section aims to provide performance metrics, inspired from \cite{Chabridon2017, Morio2015}, to assess IS-AL performances, in comparison with the RS and MLE strategies, on test cases.}

\subsubsection{\cy{Performance metrics based on the training errors \label{sec:TrainE}}}
\cy{For the IS-AL strategy, the training error is called $ \widehat{R}^{\rm IA}_{n} = \widehat{R}^{\rm IA}_{n,{\rm reg}}(\widehat{\theta}^{\rm IA}_n; \beta^{\rm IA}_{\rm reg})$  and is defined by equation~(\ref{eq:RIA}). 

For the RS and MLE strategies, the training errors are respectively called $\widehat{R}^{\rm RS}_{n} = \widehat{R}^{\rm RS}_{n,{\rm reg}}(\widehat{\theta}^{\rm RS}_n; \beta^{\rm RS}_{\rm reg})$ and $\widehat{R}^{\rm MLE}_{n} = \widehat{R}^{\rm MLE}_{n,{\rm reg}}(\widehat{\theta}^{\rm MLE}_n; \beta^{\rm RS}_{\rm reg})$, and are defined by :
$$
\widehat{R}^{\rm \bullet}_{n,{\rm reg}}(\theta;{\beta}_{\rm reg}^{\rm RS})=  \frac{1}{n} \sum\limits_{i=1}^{n} \ell_{\theta}(X_i, S_i) + \frac{\beta_{\rm reg}^{\rm RS}}{n \beta}
$$
where $\bullet$ is for RS or MLE. Note that for MLE the penalization $\widehat{\beta}^{\rm RS}_{\rm reg}$ is only used to define similar training errors as for IS-AL and RS algorithms, in order to compare the same quantity.}

Thus, the performance metrics are :
\begin{itemize}
    \item the \textit{\R{Relative Standard Deviation}}
    \cy{
    \begin{align}\label{eq: RSD}
    {\text{RSD}}_n^{\rm \bullet} 
    = \frac{\sqrt{\mathbb{V}[\widehat{R}^{\rm \bullet}_{n}]}}{\mathbb{E}[\widehat{R}^{\rm \bullet}_{n}]} , 
    \end{align}
    where $\bullet$ is for IA, RS and MLE.}
\end{itemize}

\begin{itemize}
    \item the \textit{Relative Bias}
    \begin{equation}\label{eq: RB}
    \cy{\text{RB}_n^{\rm \bullet}  = \frac{|{\rm b} - \mathbb{E}[\widehat{R}_{n}^{\rm \bullet}]|}{\rm b}},
    \end{equation}
    \cy{where $\bullet$ is for IA, RS and MLE, and ${\rm b} = \mathbb{E}[\mu(X)(1 - \mu(X))]$.}
\end{itemize}

\begin{itemize}
    \item The \textit{efficiency} 
    \begin{equation}\label{eq: eff}
    \cy{\nu_n^{\rm \bullet} = \frac{\mathbb{V}[\widehat{R}_{n}^{\rm \bullet}]}{\mathbb{V}[\widehat{R}^{\rm IA}_{n}]},}
    \end{equation} 
    \cy{where $\bullet$ is for RS and MLE. A value of $\nu_n^{\rm \bullet} > 1$ shows that IS-AL has a smaller loss variance than RS or MLE}.     
\end{itemize}

\cy{The above metrics are empirically calculated using $R$ replications of the three procedures (IS-AL, RS and MLE) and $\rm b$ is estimated using a massive Monte Carlo estimator of $\mu$. In practice, this is only possible when the numerical models are not numerically expensive. This is the case for the first two numerical test cases addressed in Section \ref{sec:NumRes}, but not for the third industrial case.}

\subsubsection{\cy{Performance metrics based on the testing errors \label{sec:TestE}}}

\cy{Additionally, we define the testing error by:
$$
\widehat{Q}_{n,{\rm reg}}^{\rm \bullet}(\theta ; \beta_{\rm reg}) = \frac{1}{n_t} \sum\limits_{i=1}^{n_t} \cy{\ell_{\theta}(X_i^{(t)}, S_i^{(t)}) + \frac{\beta_{\rm reg}}{n_t \beta}} \ ,
$$
where $(X_i^{(t)}, S_i^{(t)})_{i = 1}^{n_t}$ is a testing set (independently and identically distributed with the original distribution $P$). The associated performance metrics are calculated by replacing $\widehat{R}_{n,{\rm reg}}^{\rm \bullet}(\theta ; \beta_{\rm reg})$ with $\widehat{Q}_{n,{\rm reg}}^{\rm \bullet}(\theta ; \beta_{\rm reg})$ (also called $\widehat{Q}_n^{\rm \bullet}$ in the following) as defined in section~\ref{sec:TrainE}.}


\subsection{\cy{Benchmark on the confidence ellipsoids IS-AL and MLE} \label{sec:ACEbench}}

\cy{This section aims to propose a procedure to evaluate the quality of the IS-AL asymptotic confidence ellipsoid, $\mathcal{E}_{n, \xi}^{\rm IA}$, compared to that of the MLE, $\mathcal{E}_{n, \xi}^{\rm MLE}$. This procedure is based on the use of $R$ replications of the IS-AL and MLE algorithms, as for the evaluation of the performance metrics. So, we first define the empirical estimator of $\mathbb{P}\left(\theta_* \in \mathcal{E}_{n, \xi}^{\rm IA} \right)$ (resp. $\mathbb{P}\left(\theta_* \in \mathcal{E}_{n, \xi}^{\rm MLE} \right)$), namely the Coverage Probability (CP), in order to numerically (i) verify the definitions of the ellipsoids and (ii) evaluate their convergences with respect to the size $n$ of the samples. Then, to quantify the effectiveness of the IS-AL strategy on reducing the variance of the estimate of the fragility curve, compared to that of the MLE, we define and compare their Confidence Ellipsoid Volumes (CEVs).

Section~\ref{sec:CE_CP_MLE} concerns the definitions of the confidence ellipsoid and the coverage probability for MLE. As the IS-AL confidence ellipsoid is defined in section~\ref{sec:ACE}, section~\ref{sec:CE_CP_ISAL} deals only with the associated CP. Finally section~\ref{sec:CEV} defines the CEVs for both procedures.}



\subsubsection{\cy{Confidence ellipsoid and coverage probability for MLE} \label{sec:CE_CP_MLE}}

\cy{In order to define the asymptotic confidence ellipsoid for the MLE and to compute the associated CP, we use (i) the asymptotic normality of the MLE estimator \cite{Bachoc2013} and (ii) the independence property of the samples that allows the use of the bootstrap method. 

We first consider $R$ replications of MLE estimator $\widehat{\theta}_n^{\rm MLE}$ for different sample size $n$ and build $B$ bootstrap samples of size $n$ for each replication in order to compute a bootstrap covariance:
\begin{equation}
    \widehat{V}^{\rm MLE}_{n,r} = \frac{1}{B} \sum\limits_{b=1}^B n(\theta_{b, r}^{{\rm MLE}, *} - \widehat{\theta}_{n, r}^{\rm MLE})(\theta_{b, r}^{{\rm MLE}, *} - \widehat{\theta}_{n, r}^{\rm MLE})^T\ ,
\end{equation}
where $\widehat{\theta}_{n, r}^{\rm MLE}$ is the MLE estimator for the $r$-th replication of size $n$ and $\theta_{b, r}^{{\rm MLE}, *}$ the bootstrap MLE estimator for the $b$-th bootstrap sample of the $r$-th replication.

Thus, the bootstrapped confidence ellipsoid for MLE is defined by:
\begin{equation}
\label{eq:bootconfidenceellipsoid}
    \mathcal{E}^{{\rm MLE},r}_{n,\xi} = \{\theta : n(\theta - \widehat{\theta}_{n, r}^{\rm MLE})^T(\widehat{V}_{n,r}^{\rm MLE})^{-1}(\theta - \widehat{\theta}_{n, r}^{\rm MLE}) \leq q_{1 - \xi}^{\chi^2(m)}\} \ ,
\end{equation}
while the bootstrap CP writes:
\begin{equation}
     {\rm CP}_{n}^{{\rm MLE},r} = \frac{1}{R} \sum\limits_{r=1}^R \mathds{1}_{\theta_* \in \mathcal{E}^{{\rm MLE},r}_{n, \xi}} \ .
\end{equation}}

\subsubsection{\cy{Coverage probability for IS-AL} \label{sec:CE_CP_ISAL}}

\cy{The IS-AL asymptotic confidence ellipsoid $\mathcal{E}_{n, \xi}^{\rm IA}$ is defined in section~\ref{sec:ACE} by equation (\ref{eq:asymp ce}). So, as for MLE, the associated CP is computed by considering $R$ replications of IS-AL, namely:
$$
{\text{CP}}_{n}^{{\rm IA},r} = \frac{1}{R} \sum\limits_{r=1}^R \mathds{1}_{\theta_* \in \mathcal{E}^{{\rm IA},r}_{n,\xi}} \ ,
$$
where $\mathcal{E}^{{\rm IA},r}_{n,\xi}$ is the asymptotic confidence ellipsoid of the $r$-th replication of the IS-AL procedure of size $n$.}

\subsubsection{\cy{Confidence ellipsoid volumes for IS-AL and MLE} \label{sec:CEV}}

\cy{A qualitative criterion to measure the sharpness of a confidence ellipsoid is its volume \cite{Golestaneh2018}. So, to evaluate the effectiveness of the IS-AL strategy on the reduction of the variance of the fragility curve estimations, we define the CEVs, for respectively the MLE and IS-AL strategies, as follows:
\begin{equation}
    \text{CEV}_n^{{\rm MLE},r} = \det\left(\frac{\widehat{V}^{\rm MLE}_{n,r}}{n}\right) \ ,
\end{equation}
and
\begin{equation}
    \text{CEV}_{n}^{{\rm IA},r} = \det\left(\frac{ \widehat{G}_{n,r} }{n}\right) ,
\end{equation}
where $\widehat{G}_{n,r}$ is the estimated covariance matrix (\ref{eq:Gn}) of the $r$-th replication of IS-AL procedure of size $n$.}


\section{\R{Numerical results}}
\label{sec:NumRes}%

To evaluate IS-AL efficiency, a numerical benchmark has been performed with three test cases with increasing complexity:\\
1) a synthetic test case with known fragility curve and probability distribution of the seismic log-intensity measure $X$, \\
2) \R{a nonlinear elasto-plastic oscillator with kinematic hardening subjected to synthetic signals generated from the modulated and filtered white-noise ground-motion model \cite{Rezaeian10}, as in \cite{Sainct20},}\\
3) \R{an industrial test case of a nuclear facility's pipeline-system, submitted to the same artificial signals.} 

\cy{For test cases 2 and 3, 97 acceleration records  selected from the European Strong Motion Database \citep{ESMD} in the domain $ 5.5 < M < 6.5$ and $R < 20 {\rm km}$ - where $M$ is the magnitude and $R$ the distance from the epicenter - are considered in order to identify the parameters of the ground-motion model. $10 ^ 5$ realizations of synthetic signals are then generated to form the unlabeled pool.}

\R{The oscillator test case aims to evaluate the effectiveness of the IS-AL strategy before its application to an industrial test case which is numerically much more costly. Moreover, since it well represents the essential features of the nonlinear responses of a large variety of real structures subjected to earthquakes, this test case allows to determine the value of the hyperparameter $\varepsilon$ - thanks to a numerical benchmark - because there is no ad hoc procedure to do this.}

\subsection{Synthetic test case}
Here we benchmark our methodology while having full knowledge of the true fragility curve. We generate 30,000 datapoints $(X_i, S_i)$ with the fragility curve $\mu(x) = \Phi(\frac{x - \log(\alpha_*)}{\beta_*})$ with $(\alpha_*, \beta_*) = (0.3, 0.4)$. The original marginal distribution of $X$ is here a Gaussian distribution with mean $\log\left(\frac{\alpha_*}{5}\right)$ and variance $1.69$.
\modif{The parameters have been chosen so that the data generated are qualitatively close to the nonlinear oscillator test case presented in section \ref{sec:osciNL}.}
The unlabeled pool consists of 20,000 datapoints $X_i$. 10,000 datapoints $(X_i, S_i)$ will be our validation set for testing error estimation, using crude Monte Carlo.

\begin{figure}[!ht]
    \centering
    \includegraphics[width=9cm]{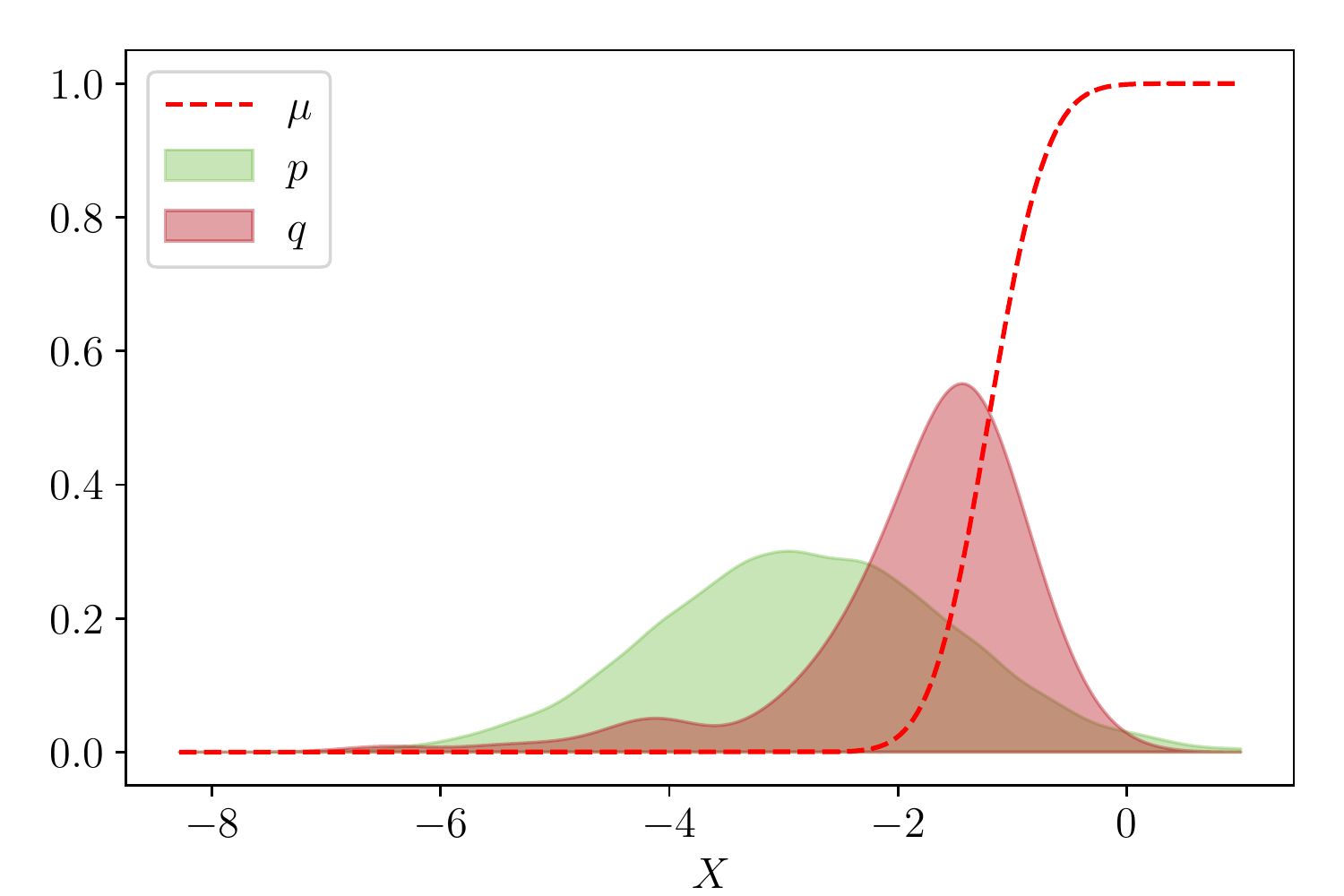}
    \caption{\R{Synthetic test case with lognormal fragility curve with parameters $(\alpha_*, \beta_*) = (0.3, 0.4)$ and $ X \sim \mathcal{N}(\frac{\alpha_*}{5}, 1.69)$. Comparison of the original marginal density $p$ of $X$ with the empirical density $q$ of the $n = 120$ datapoints $X_i$ obtained by IS-AL.}}
    \label{fig:synthetic case marginal}
\end{figure}

Figure \ref{fig:synthetic case marginal} shows (i) the target fragility curve $\mu$ in dashed red line, (ii) a kernel density estimation of the density $p$ based on the whole dataset in green and (iii) a kernel density density estimation $q$ of the 120 datapoints $X_i$ obtained by IS-AL in red. 

\begin{figure}[!ht]
     \centering
     \begin{subfigure}[b]{0.48\textwidth}
         \centering
         \includegraphics[width=\textwidth]{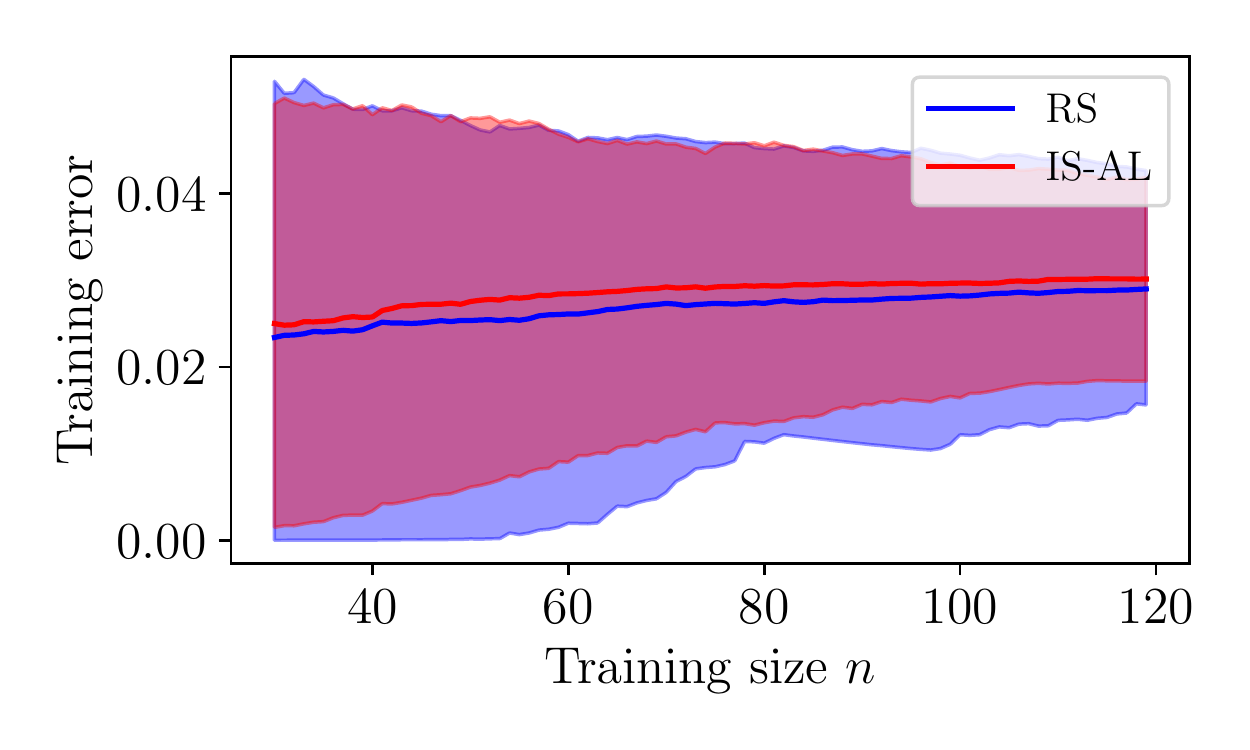}
         \caption{$\widehat{R}^{\rm IA}_{n}$ versus $\widehat{R}^{\rm RS}_{n}$}
         \label{fig:synth train}
     \end{subfigure}
     \hfill
     \begin{subfigure}[b]{0.48\textwidth}
         \centering
         \includegraphics[width=\textwidth]{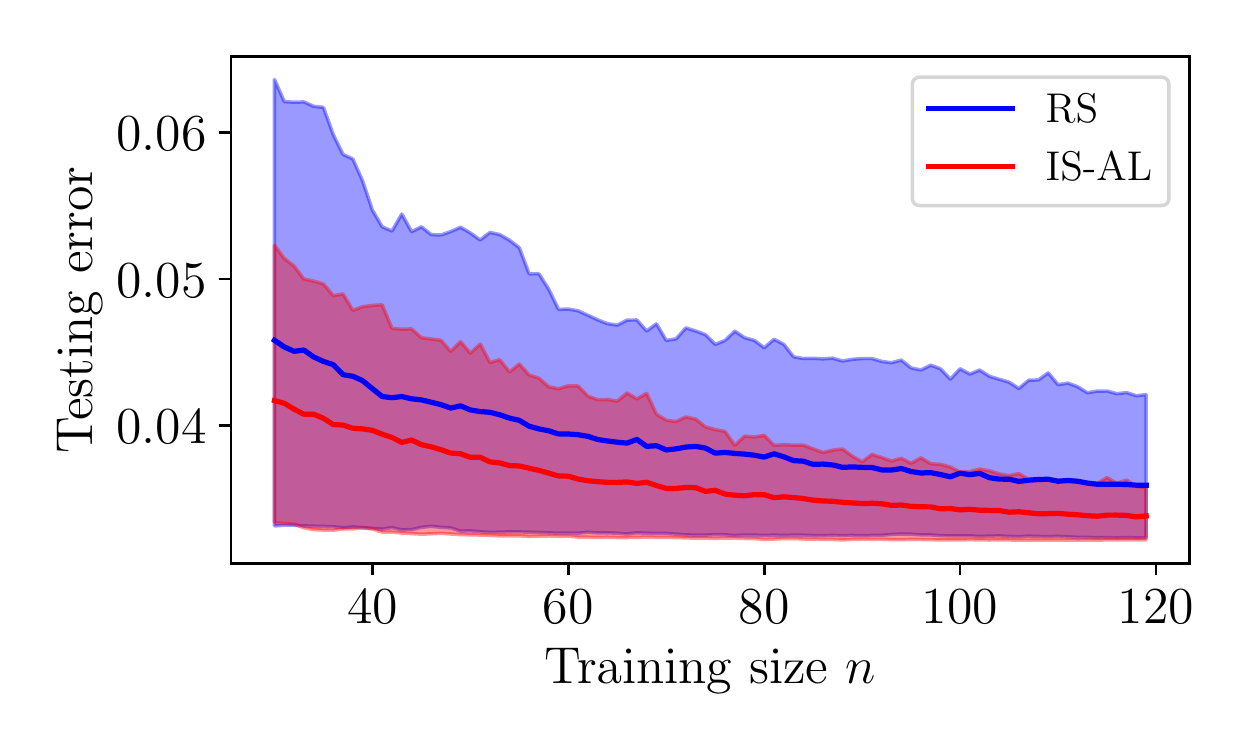}
         \caption{$\widehat{Q}^{\rm IA}_{n}$ versus $\widehat{Q}^{\rm RS}_{n}$}         
         \label{fig:synth test}
     \end{subfigure}
     \begin{subfigure}[b]{0.48\textwidth}
         \centering
         \includegraphics[width=\textwidth]{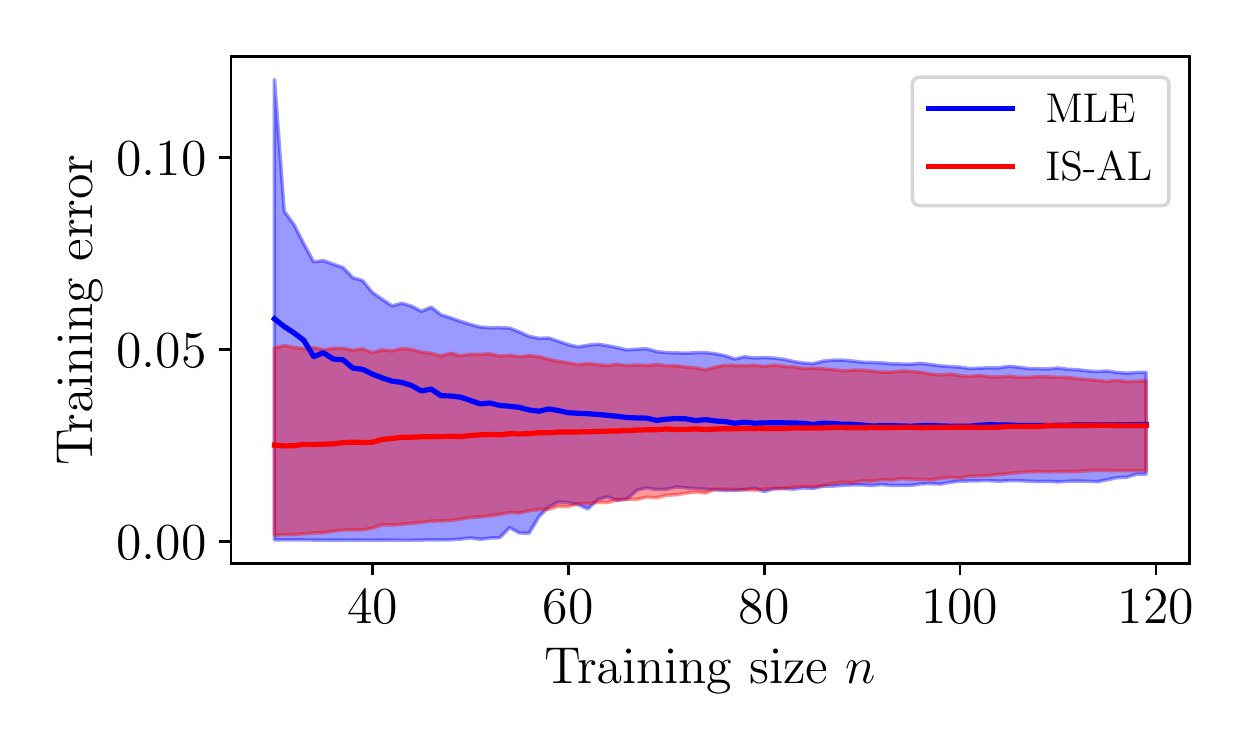}
         \caption{$\widehat{R}^{\rm IA}_{n}$ versus $\widehat{R}^{\rm MLE}_{n}$}
     \end{subfigure}
     \hfill
     \begin{subfigure}[b]{0.48\textwidth}
         \centering
         \includegraphics[width=\textwidth]{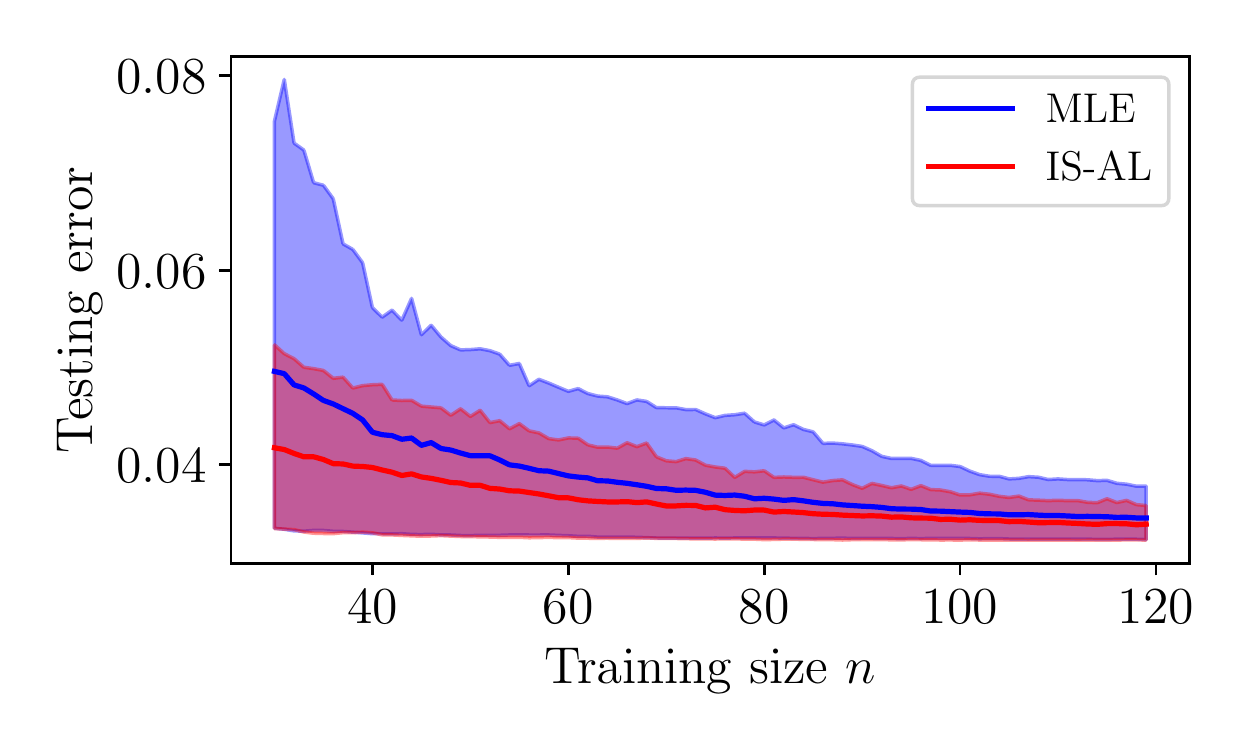}
         \caption{$\widehat{Q}^{\rm IA}_{n}$ versus $\widehat{Q}^{\rm MLE}_{n}$}            
     \end{subfigure}
     \caption{Results of the numerical benchmark for the synthetic test case: the thick lines represent the mean loss for $R = 500$ replications, the shaded areas represent the ranges between the quantiles at $90\%$ and $10\%$ of the $500$ replications of the IS-AL, RS and MLE procedures. \R{For this case, the bias is known and is equal to $\mathbb{E}[ \mu(X)(1-\mu(X))] \simeq 0.032$.}}
     \label{fig:synth loss}
\end{figure}

\cy{Figure \ref{fig:synth loss} shows the training and testing errors for $R = 500$ replications of the IS-AL, RS and MLE algorithms. The algorithms are initialized with $20$ datapoints and $n = 100$ datapoints are extracted from the unlabeled pool with the three procedures. The regularization parameters $\beta_{\rm reg} \in (10^{-4}, 10^{-1})$ were determined by cross validation with the $20$ datapoints used for initialization for each replication of the IS-AL and MLE strategies. We also use a defensive parameter value $\varepsilon$ of $10^{-3}$ (see section~\ref{sec:osciNL} for justification).}


\begin{table}[!ht]
    \caption{Performance metrics for the synthetic test case for $n = 120$}
    \label{table:synth}
    \begin{center}
        \begin{tabular}{ c|c|c|c||c|c|c } 
         \multicolumn{1}{c}{} & \multicolumn{3}{c}{Train} & \multicolumn{3}{c}{Test} \\
         \hline
         {${\rm \bullet}$} & RS & MLE & IS-AL & RS & MLE & IS-AL \\ 
         \hline
         $\text{RSD}_{120}^{\rm \bullet}$ ($\%$) & $38$ & $36$ & $34$ & $12$ & $9$ & $8.5$ \\
         $\nu_{120}^{\rm \bullet}$ & $1.2$ & $1.2$ & $\times$ & $2.3$ & $1.1$ & $\times$ \\
         $\text{RB}_{120}^{\rm \bullet}$ ($\%$) & $8.6$ & $4.2$ & $5$ & $13$ & $9$ & $6.7$ 
        \end{tabular}
    \end{center}
\end{table}

As depicted by Figure~\ref{fig:synth loss} and Table \ref{table:synth}, IS-AL does not seem to reduce the training error. This result is normal because IS-AL selects seisms whose intensity measures maximize $\tilde{\ell}_{\theta}$, which can be seen as a marginalized training loss variance of the observations. \R{In other words, as illustrated in Figure \ref{fig:synthetic case marginal} with the density $q(x)$, IS-AL selects "difficult" points - typically values of $ x $ for which $ \mu(x) $ takes values between $0$ and $1$ -} and therefore the training error can be large because it is not representative of the generalization error as the testing one. RS, MLE and IS-AL strategies really distinguish themselves on the testing error, which is smaller for IS-AL. Moreover, IS-AL quickly converges to the known bias equal to $\mathbb{E}[ \mu(X)(1-\mu(X))] \simeq 0.032$. In comparison with RS and MLE strategies, the variance of IS-AL is smaller after $120$ iterations: $\nu_{120}^{\text{MLE}}$ is smaller than $\nu_{120}^{\text{RS}}$, meaning that MLE is competitive with IS-AL in this synthetic case.

\begin{figure}[!ht]
     \centering
     \begin{subfigure}[b]{0.48\textwidth}
         \centering
         \includegraphics[width=\textwidth]{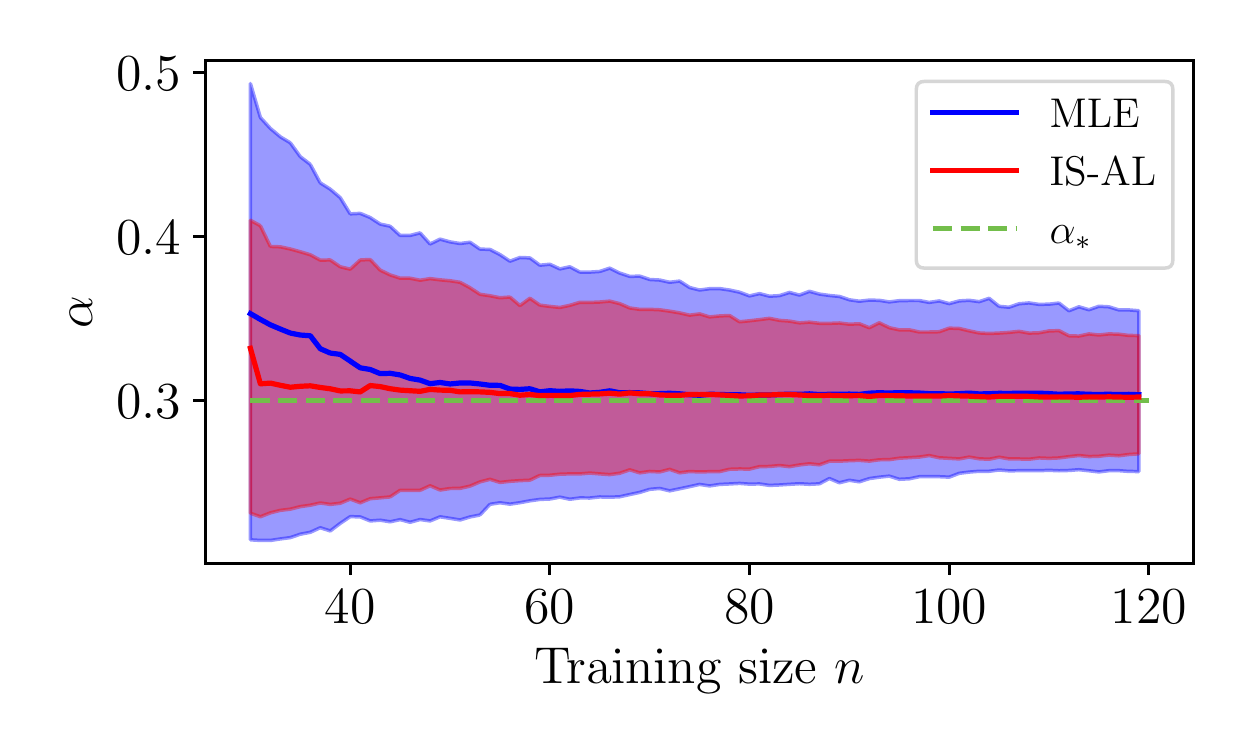}
     \end{subfigure}
     \hfill
     \begin{subfigure}[b]{0.48\textwidth}
         \centering
         \includegraphics[width=\textwidth]{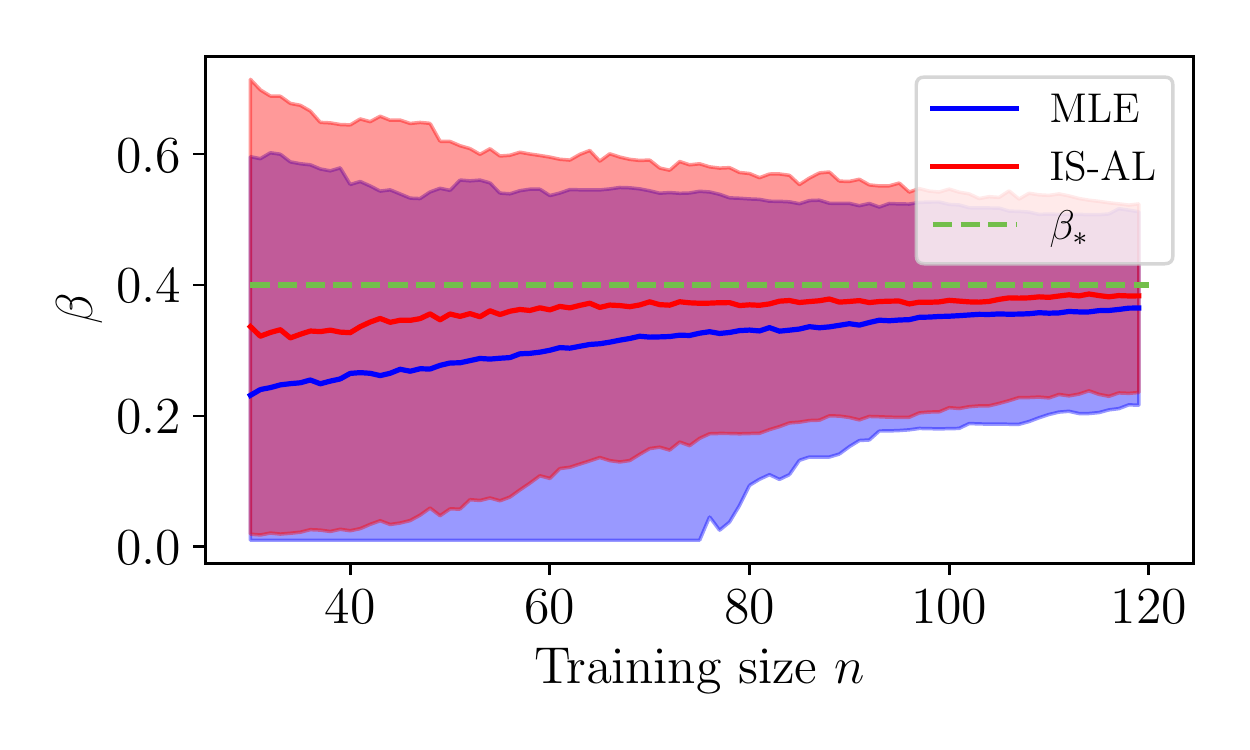}
     \end{subfigure}
     \caption{\modif{Results of the numerical benchmark for the synthetic test case: empirical distributions of the parameters $\alpha$ and $\beta$ are represented by ranges between the empirical $10\%$ and $90\%$ quantiles of $500$ replications. The shaded blue and red areas correspond respectively to MLE and IS-AL. The dashed green lines correspond to the true parameters $\alpha_*$ and $\beta_*$.}}
     \label{fig:synth mle params}
\end{figure}

Figure \ref{fig:synth mle params} compares the distributions of the parameters $\alpha$ and $\beta$ for several sample sizes using the $500$ replications of MLE and IS-AL estimators and they are similar when $n > 100$. Indeed, the statistical model is in this case well specified (i.e. failure events follow a Bernoulli distribution with a lognormal probability of failure) and thus MLE is supposed to perform well as shown in \cite{Bachoc2013}. Note that up to $n = 80$, the MLE strategy can produce degenerate fragility curves because $\beta \simeq 0$. The IS-AL algorithm avoids this pitfall due to the regularization parameter. 


\subsection{A nonlinear oscillator}\label{sec:osciNL}

\cy{This test case aims to validate the overall strategy developed in this work on a simple but representative case, because this is not possible for complex structures like the one in section \ref{sec:industestcase}. This section is therefore particularly comprehensive, from the initialization of the IS-AL algorithm to the estimations of the fragility curves, via the choice $\varepsilon$ and the numerical verification of the theorems.}

\subsubsection{Presentation of the oscillator}

\begin{figure}[!ht]
    \centering
    \includegraphics[width=5cm]{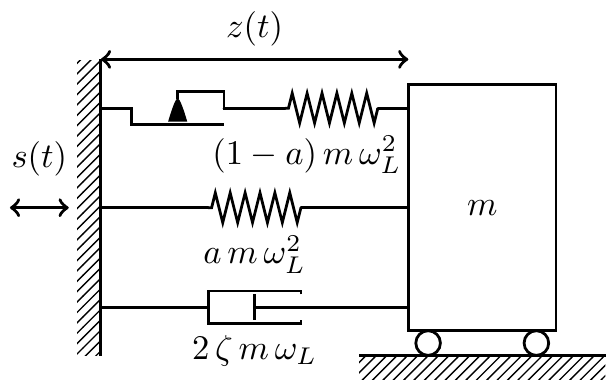}
    \caption{Elasto-plastic mechanical oscillator with kinematic hardening, with parameters $f_L = 5$ Hz and $\zeta = 2\%$. The yield limit is $Y = 5.10^{-3}$ m and the post-yield stiffness is $20\%$ of the elastic stiffness, hence $a = 0.2$.}
    \label{fig:KH}
\end{figure}

This second test case - illustrated in Figure \ref{fig:KH} - relates to a single degree of freedom elasto-plastic oscillator which exhibits kinematic hardening. It has been used in previous studies such as~\cite{TREVLOPOULOS2019, Sainct20}. For a unit mass $m$, its equation of motion is:
$$
\ddot{z}(t) + 2 \zeta \omega_L\dot{z}(t) + f_{NL}(t) = -s(t) \ ,
$$
with $s(t)$ an artificial seismic signal. $\dot{z}(t)$ and $\ddot{z}(t)$ are respectively the velocity and the acceleration of the mass while $\zeta$ is the damping ratio and $\omega_L$ the pulsation of the oscillator. The nonlinear force $f_{NL}$ is governed by two parameters: the post-yield stiffness, $a$, and the yield displacement, $Y$.

\cy{With this model, the quantity of interest is the maximum displacement of the mass, $D = \max_{t\in [0, T]}|z(t)|$, where $T$ is the duration of the seismic excitation. The failure state is then defined by the $\{0,1\}$-valued variable $S =\mathds{1}_{(D > C)}$, where $C=2Y$ is chosen \modif{to be approximately the $90\%$ quantile of the maximal linear displacement of the unlabeled pool of size $10^5$}.}

In order to check the performances of the IS-AL algorithm, the unlabeled training set consists in $9.10^4$ seismic signals and the testing set is composed of $10^4$ signals. The benchmark study consists in $R = 500$ replications with $n = 120$ sampled seismic signals using IS-AL (that includes the initial $20$ points) and $120$ for the RS and MLE strategies.

\subsubsection{Initialization of the IS-AL procedure}

\begin{figure}[!ht]
    \centering
    \includegraphics[width=9cm]{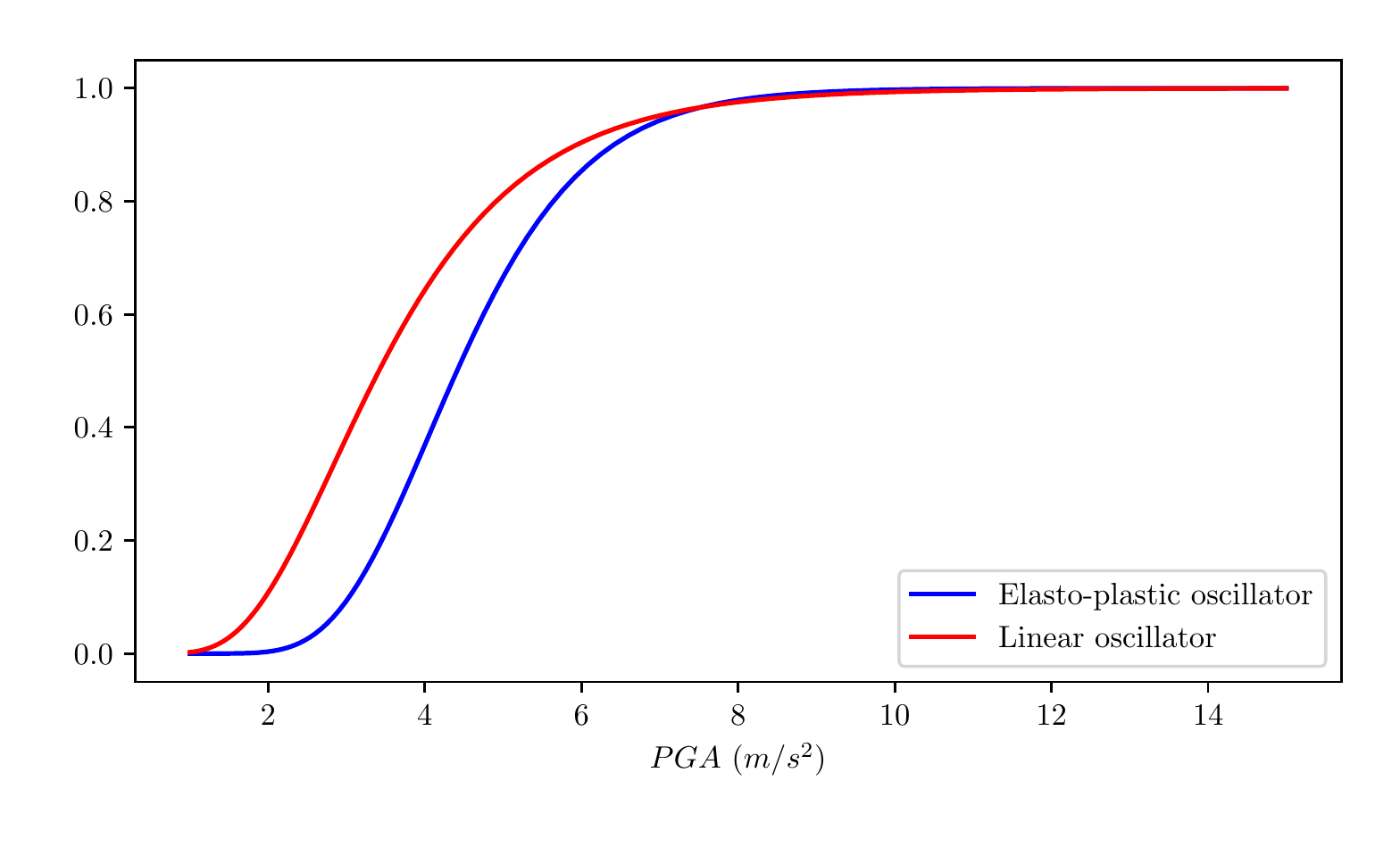}
    \caption{\cy{Lognormal fragility curves of the linear elastic and the nonlinear oscillators obtained by using least squares minimization on the total $10^5$ synthetic seismic signals of the dataset.}}
    \label{fig:linear_KH}
\end{figure}

In this test case, for IS-AL initialization, we use the underlying elastic oscillator as a cheap model. The initialization parameter $\modif{\widehat{\theta}_0^{\rm IA}}$ is approximated by $\widehat{\theta}^{\modif{\rm RS}}_{10^5}$ (equation~(\ref{R_RS})) using the $10^5$-sized dataset. In addition, the PGA is first considered as IM. Even if the PGA is not known to be the best indicator, doing so helps to verify the relevance of the methodology in a "less favorable" case. Note that the influence of the IM on the results is discussed in section~\ref{sec:empFC_inflIM}. As shown in Figure \ref{fig:linear_KH}, the parameter $\modif{\widehat{\theta}_0^{\rm IA}}$ could be considered "close to" the true parameter $\theta_*$. Thus, $20$ datapoints are queried on the nonlinear oscillator with the instrumental density $q_{\modif{\widehat{\theta}}^{\rm IA}_0,\varepsilon}$ (equation~(\ref{eq:qteps}))  before launching the adaptive strategy.

\subsubsection{Choice of $\varepsilon$}

\cy{As mentioned in section~\ref{sec:ISALbeta}, there is no direct methodology for the choice of the $\varepsilon$ value. One thus benefits from this simple test case to implement a numerical benchmark in order to obtain a reasonable value of $\varepsilon$ for the class of structures for which the oscillator represents the global nonlinear behavior under seismic excitation. This benchmark consists in evaluating the IS-AL efficiency with respect to the RS strategy, $\nu_n^{\rm RS}$ (equation~(\ref{eq: eff})), as a function of $\varepsilon$ when IM = PGA and $n = 120$. Results are given in Table~\ref{table: defensive_benchmark}. 

\begin{table}[!ht]
    \caption{\cy{Defensive parameter $\varepsilon$ influence on $\nu_{120}^{\rm RS}$ when IM = PGA.}}
    \label{table: defensive_benchmark}
    \begin{center}
            \begin{tabular}{c|c|c}
                 $\varepsilon$ & Train & Test  \\
                 \hline
                 $10^{-1}$ & $1.3$ & $1.2$ \\
                 $10^{-2}$ & $2.1$ & $3.9$ \\
                 $10^{-3}$ & $2.2$ & $3.3$
            \end{tabular}
    \end{center}
\end{table}

They show that $\nu_{120}^{\rm RS}$ does not change between $\varepsilon = 10^{-2}$ and $\varepsilon = 10^{-3}$. $\nu_{120}^{\rm RS}$ is smaller when $\varepsilon = 10^{-1}$ meaning that this value is too conservative because there are too many elements drawn from the pdf $p$. Accordingly, all the results will presented hereafter with a defensive parameter $\varepsilon = 10^{-3}$.
This implies that the defensive strategy plays essentially no role here, but gives theoretical convergence guarantees.}

\subsubsection{Performance metrics}

\begin{figure}[!ht]
     \centering
     \begin{subfigure}[b]{0.48\textwidth}
         \centering
         \includegraphics[width=\textwidth]{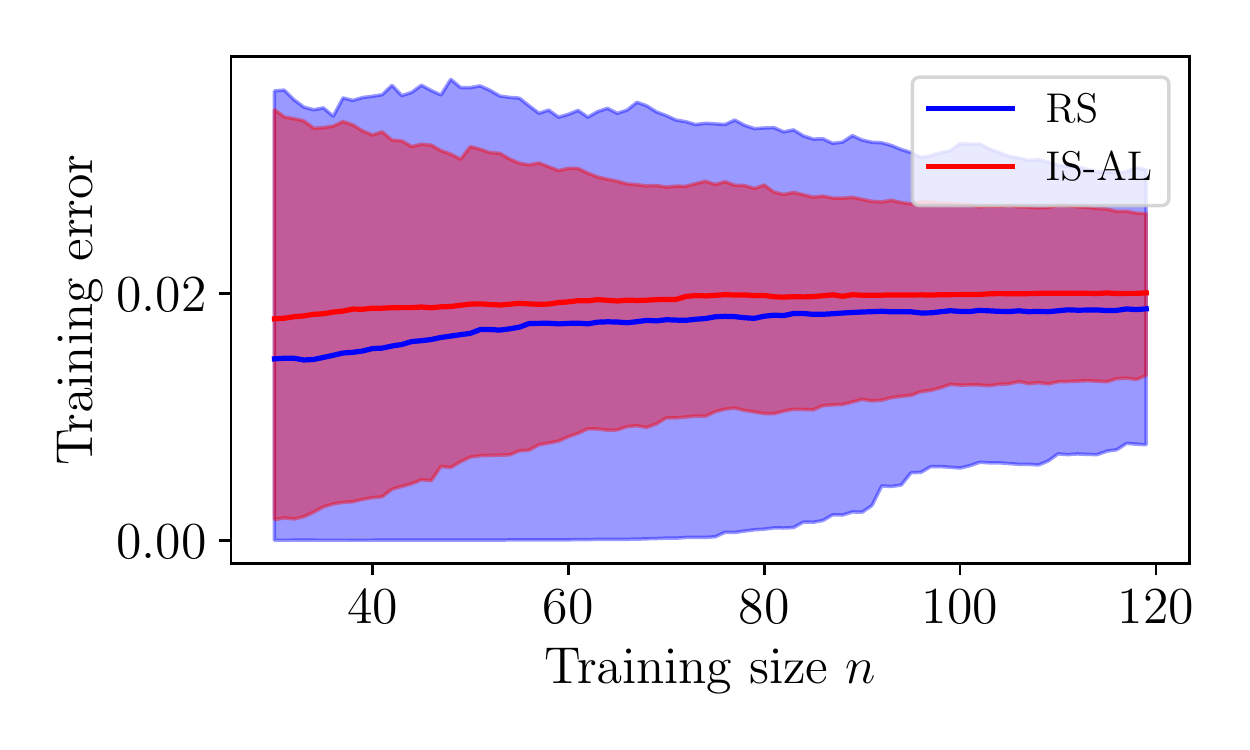}
         \caption{$\widehat{R}^{\rm IA}_{n}$ versus $\widehat{R}^{\rm RS}_{n}$}         
         \label{fig:kh train}
     \end{subfigure}
     \hfill
     \begin{subfigure}[b]{0.48\textwidth}
         \centering
         \includegraphics[width=\textwidth]{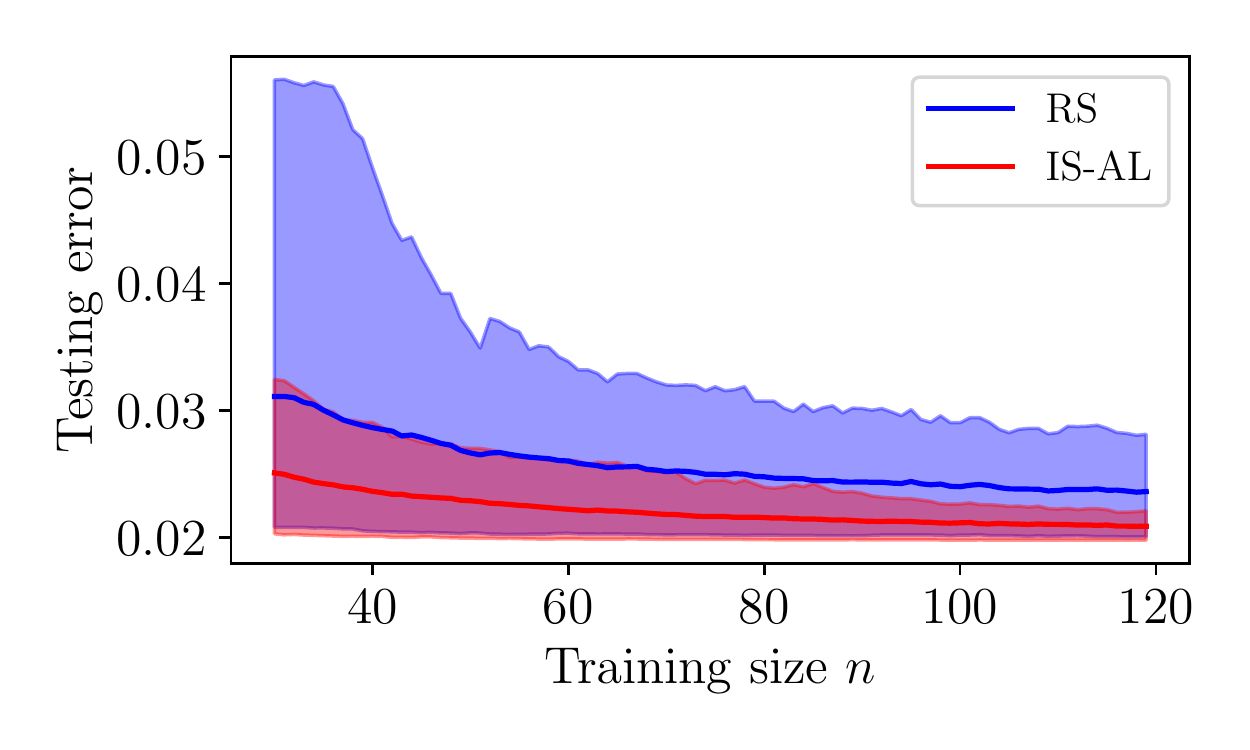}
         \caption{$\widehat{Q}^{\rm IA}_{n}$ versus $\widehat{Q}^{\rm RS}_{n}$}              
         \label{fig:kh test}
     \end{subfigure}
     \begin{subfigure}[b]{0.48\textwidth}
         \centering
         \includegraphics[width=\textwidth]{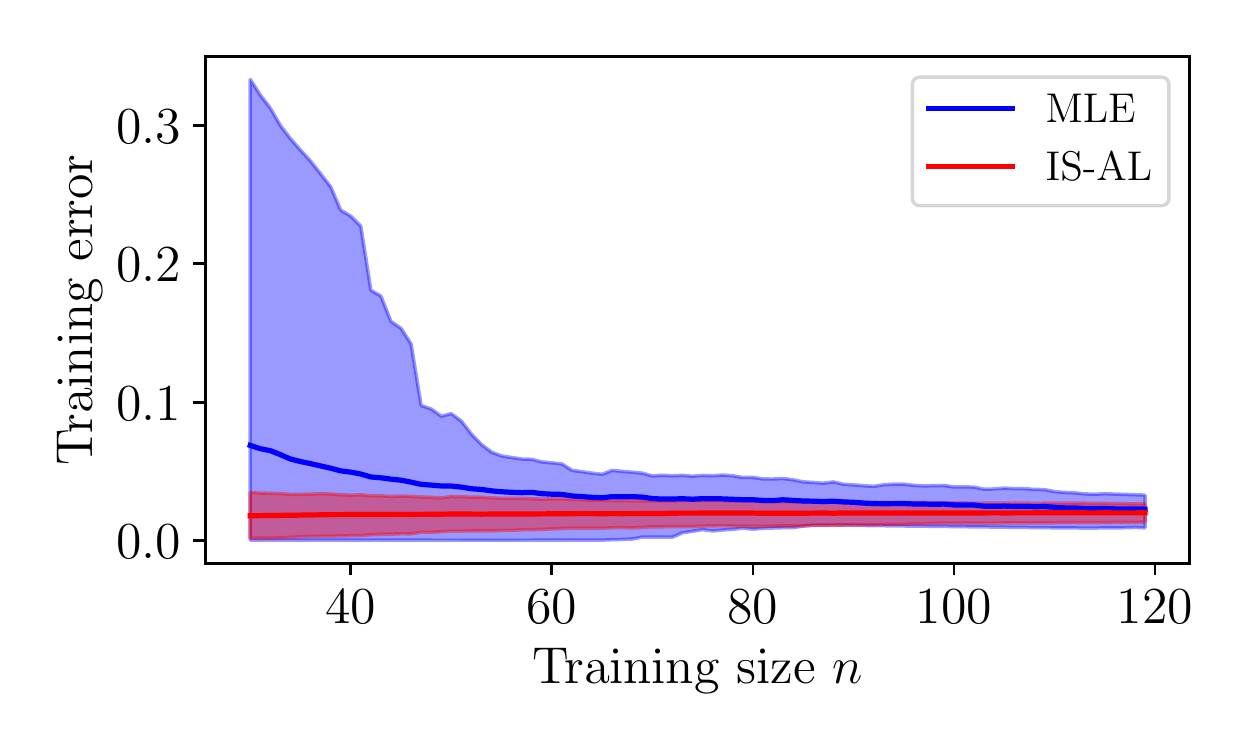}
         \caption{$\widehat{R}^{\rm IA}_{n}$ versus $\widehat{R}^{\rm MLE}_{n}$}         
         \label{fig:kh mle train}
     \end{subfigure}
     \hfill
     \begin{subfigure}[b]{0.48\textwidth}
         \centering
         \includegraphics[width=\textwidth]{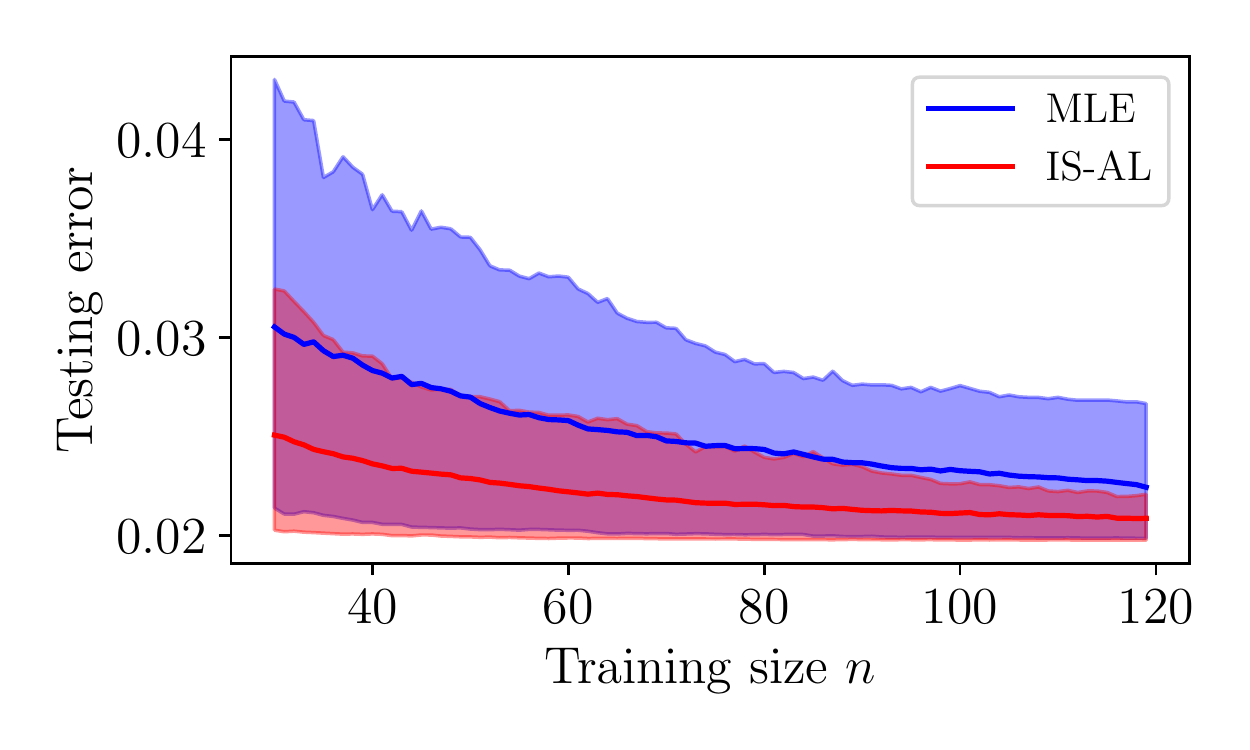}
         \caption{$\widehat{Q}^{\rm IA}_{n}$ versus $\widehat{Q}^{\rm MLE}_{n}$}         
         \label{fig:kh mle test}
     \end{subfigure}
     \caption{Results of the numerical benchmark for the elasto-plastic mechanical oscillator with the PGA as IM (same notations as for the synthetic test case). The empirical distributions of the training and testing errors are represented by the range between the empirical $90\%$ and $10\%$ quantiles of the $500$ replications.}
     \label{fig:kh loss}
\end{figure}

\cy{Figure \ref{fig:kh loss} compares the IS-AL, MLE and RS training and testing errors as functions of $n$. The mean training loss for the $500$ replications is higher for IS-AL than for RS. Indeed, the instrumental density is chosen to sample seismic signals that maximize the loss variance, resulting in a high training error. Moreover, the mean testing error of IS-AL is also significantly smaller than for RS and quickly converges to the "minimal" error related to the term $\mathbb{E}[ \mu(X)(1-\mu(X))]$ in (\ref{eq:expanderror}). This is shown in Table \ref{table:KH} by a significantly smaller value of relative bias $\text{RB}_{120}^{\rm IA}$ ($1\%$) for the testing error than with RS ($12\%$). With respect to the MLE, one cannot make equivalent remarks insofar as the two errors are "artificial" (see definitions in section~\ref{sec:TrainE}) and only plotted for illustration purpose. However, Table \ref{table:KH} shows that the IS-AL strategy has overall better performance than the other two strategies.}


\begin{table}[!ht]
    \caption{Performance metrics for the elasto-plastic oscillator for $n = 120$ when IM = PGA.}
    \label{table:KH}
    \begin{center}
        \begin{tabular}{ c|c|c|c||c|c|c } 
         \multicolumn{1}{c}{} & \multicolumn{3}{c}{Train} & \multicolumn{3}{c}{Test} \\
         \hline
         {${\rm \bullet}$} & RS & MLE & {IS-AL} & RS & MLE & {IS-AL} \\ 
         \hline
          {$\text{RSD}_{120}^{\rm \bullet}$} ($\%$) & $47$ & $65$ & $30$ & $19$ & $14$ & $11$ \\
          {$\nu_{120}^{\rm \bullet}$} & $2.2$ & $5.9$ & $\times$ & $3.3$ & $1.8$ & $\times$ \\
          {$\text{RB}_{120}^{\rm \bullet}$} ($\%$) & $11.1$ & $7.6$ & $5$ & $12$ & $6.4$ & $1$
        \end{tabular}
    \end{center}
\end{table}

\subsubsection{Empirical distributions of the parameters $\alpha$ and $\beta$}

\begin{figure}[!ht]
     \centering
     \begin{subfigure}[b]{0.48\textwidth}
         \centering
         \includegraphics[width=\textwidth]{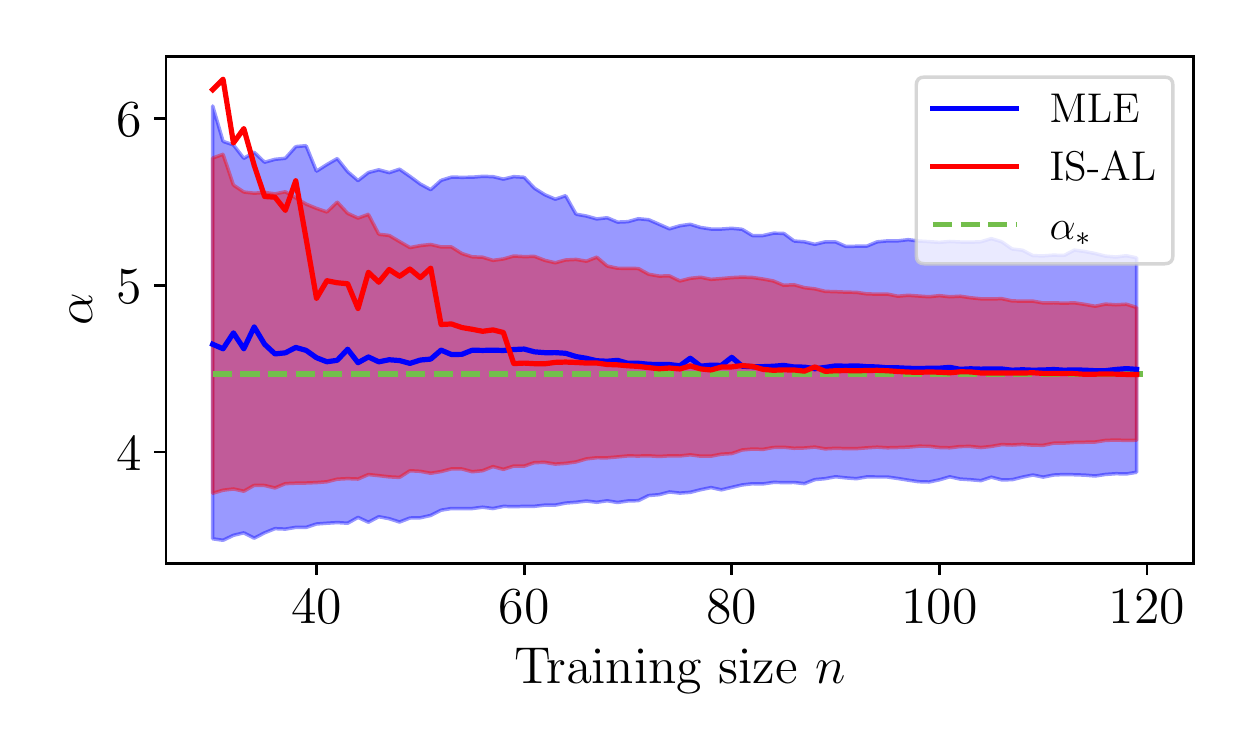}
     \end{subfigure}
     \hfill
     \begin{subfigure}[b]{0.48\textwidth}
         \centering
         \includegraphics[width=\textwidth]{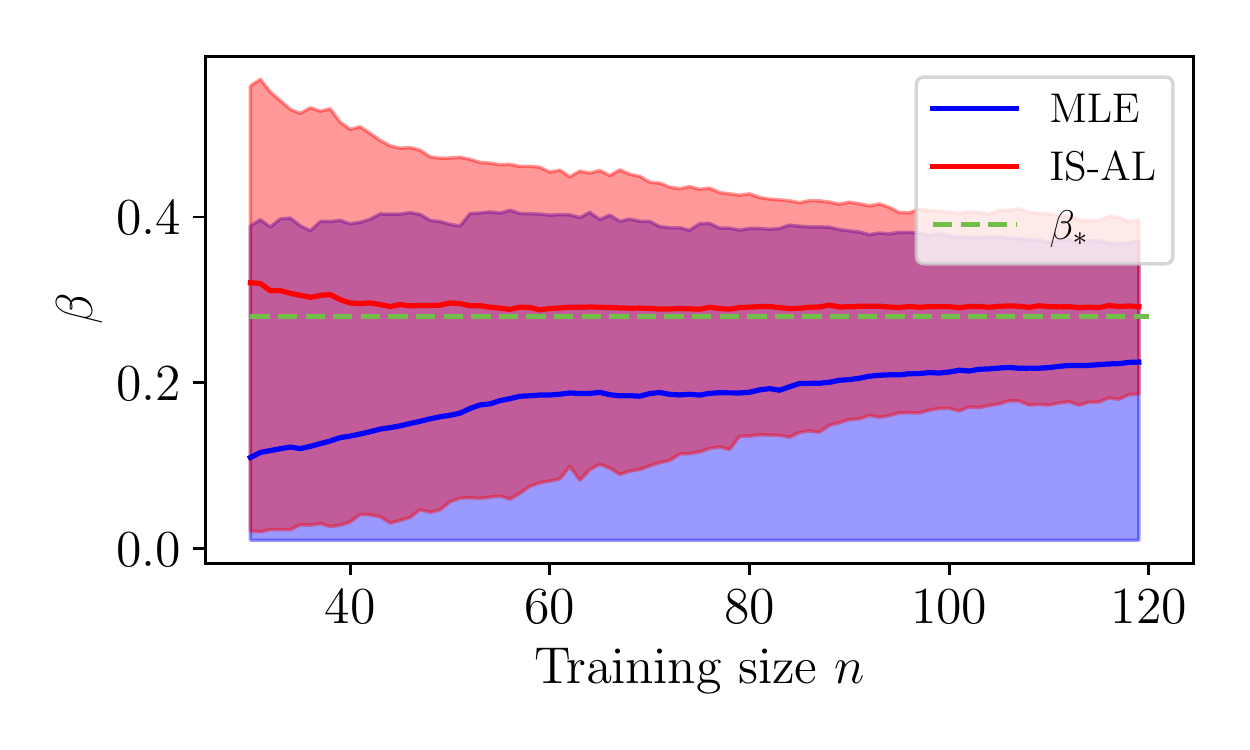}
     \end{subfigure}
     \caption{\modif{Results for the elasto-plastic mechanical oscillator with the PGA as IM: the empirical distributions of the parameters $\alpha$ and $\beta$ are represented by the empirical $90\%$ and $10\%$ quantiles of the $500$ replications and correspond to the shaded blue and red areas for MLE and IS-AL, respectively. The dashed green lines correspond to the values $\alpha_*$ and $\beta_*$, which have been here approximated by $\widehat{\alpha}_N, \ \widehat{\beta}_N$ for $N = 10^5$.}}
     \label{fig:kh mle params}
\end{figure}

\modif{Figure \ref{fig:kh mle params} shows the empirical distributions of the parameters $\alpha$ and $\beta$ for several sample sizes using $500$ replications of MLE and IS-AL estimators. Remark in this case that IS-AL performs better than the MLE by reducing the variances of the parameters' estimators. The effects are particularly visible for the parameter $\beta$, when the active learning strategy and the regularization play their role in reducing the standard deviations of the estimators without increasing bias. Indeed, MLE performances are downgraded when the model is not well specified \cite{Bachoc2013}. We remark that parameter estimation is quite unstable for IS-AL for low sample sizes. Indeed, the number of failure events for low sample sizes is often $0$, which makes impossible a correct estimation of the fragility curve's parameters.}

\begin{figure}[!ht]
    \centering
    \includegraphics[width=9cm]{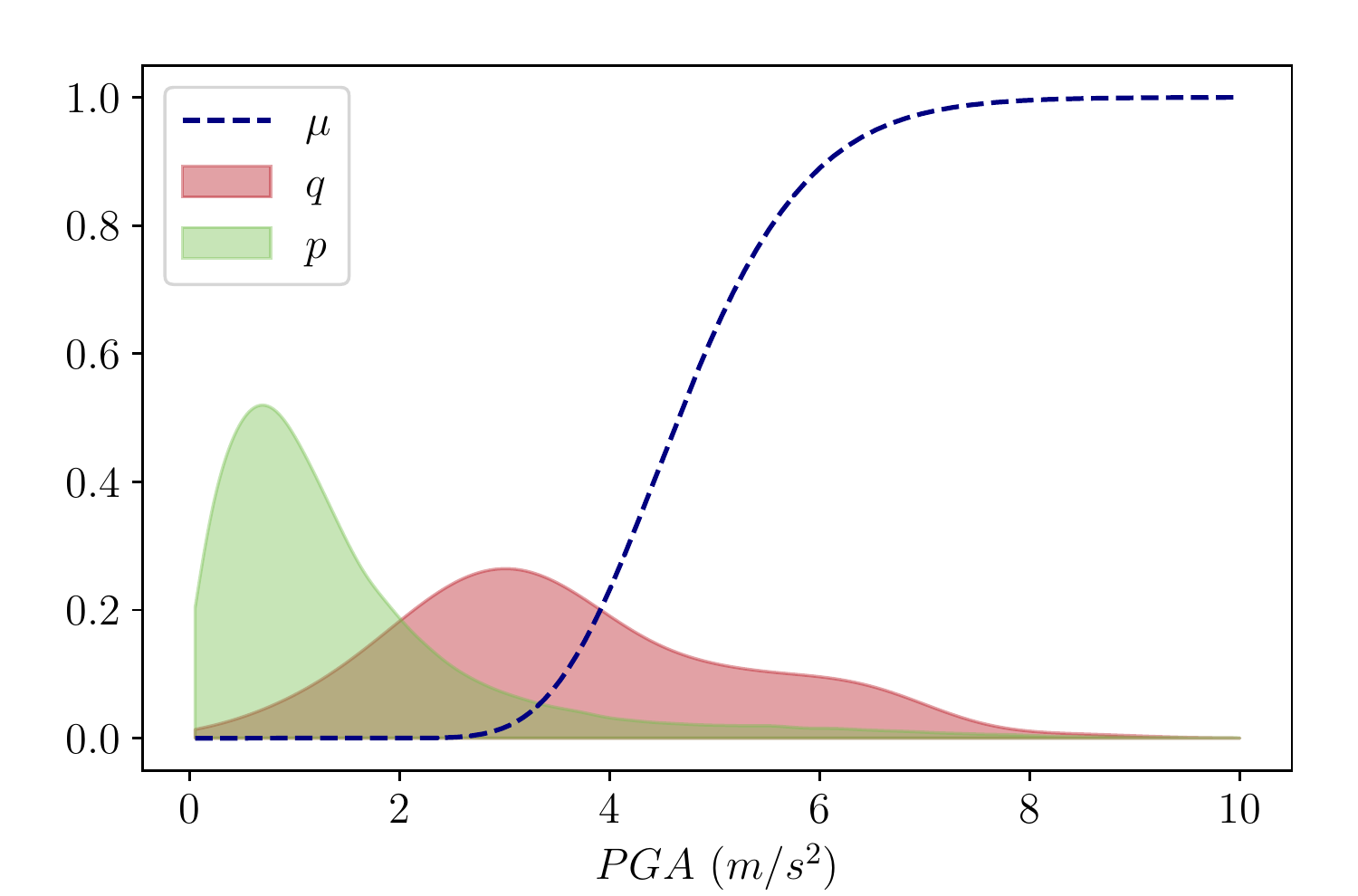}
    \caption{\cy{Comparison of the original marginal density $p$ of $PGA$ with the estimated density $q$ of sampled $PGA$ using IS-AL with $n=100$ datapoints for the nonlinear oscillator. Fragility curve is approximated by $ \mu({\rm PGA}) = \Phi( \frac{ \log( {\rm PGA} /\widehat{\alpha}_N) }{ \widehat{\beta}_N} )$ for $N = 10^5$.}}
    \label{fig:kh_marginal}
\end{figure}

Figure \ref{fig:kh_marginal} helps to visualize how IS-AL reduces the uncertainty of the fragility curve estimation: IS-AL is designed to sample seismic ground motions in the transition zone between $0$ and $1$ of the fragility curve, this phenomenon is responsible for the uncertainty reduction.


\subsubsection{Convergence criterion}

\begin{figure}[!ht]
    \centering
    \includegraphics[width=9cm]{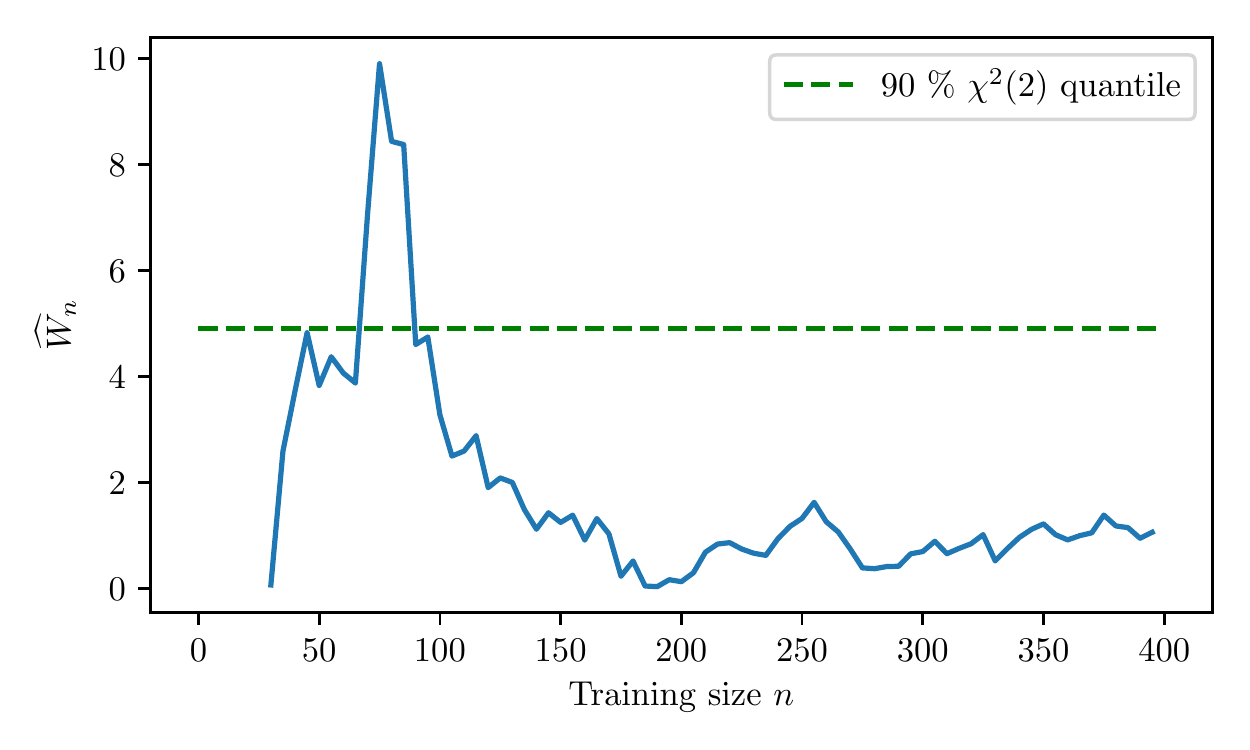}
    \caption{Values of the test statistic $\widehat{W}_n$ for two independent IS-AL realizations, when~IM~=~PGA.}
    \label{fig:stop_crit}
\end{figure}

\cy{Figure \ref{fig:stop_crit} shows the value of the test statistics $\widehat{W_n}$ (see section~\ref{sec:convergence crit}) for two independent IS-AL realizations. This result expresses that the IS-AL algorithm achieves asymptotic normality from $n = 100$ because the value of $\widehat {W} _n$ is less than the quantile $90\%$ of the distribution $\chi^2(2)$.}

\newpage

\subsubsection{CPs and CEVs \label{sec:emp_CP_CEV}}

\modif{Figure \ref{fig:pic plot} shows the CP values for the nonlinear oscillator for a training size $n$ between $100$ and $500$ for the fragility curve estimation by MLE or IS-AL. The true parameter $\theta_*$ for this case has been approximated by $\widehat{\theta}_N$ for $N=10^5$. The numerical results show that the CP values are close to the theoretical and nominal value of $90\%$, which validates the theoretical results of the section~\ref{sec:thres}.

\begin{figure}[!ht]
    \centering
    \includegraphics[width=9cm]{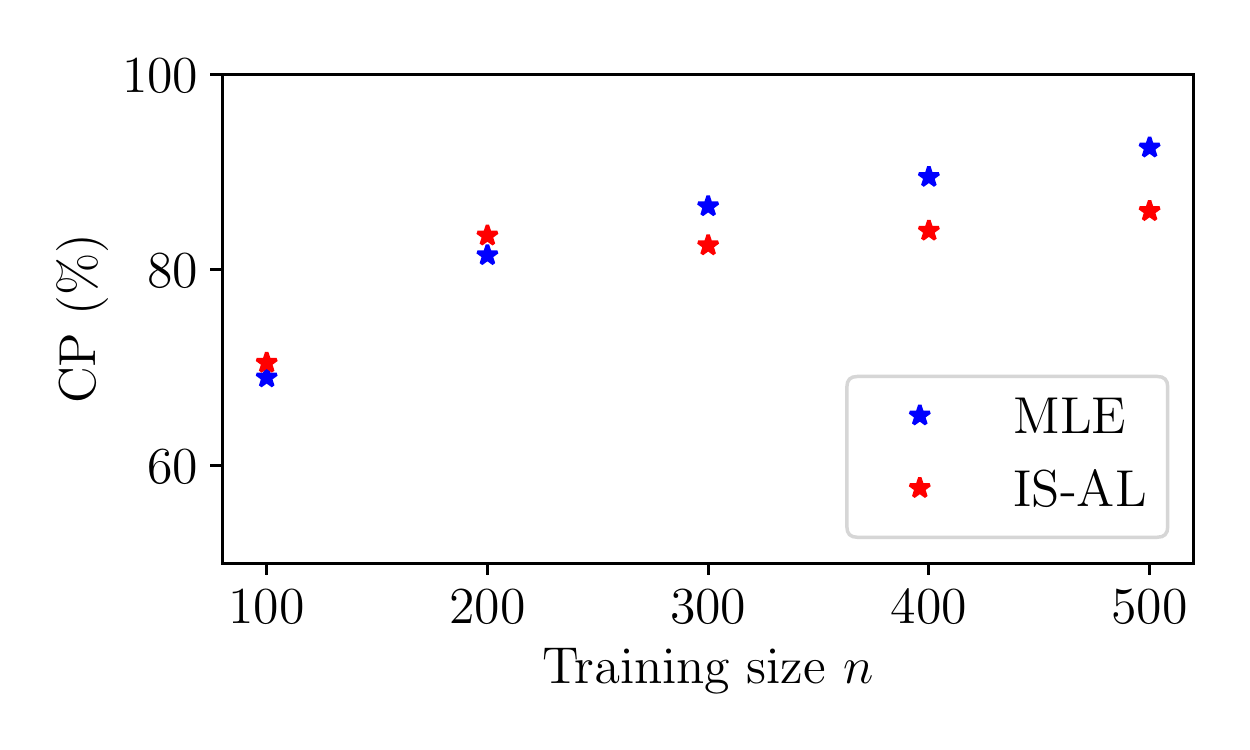}
    \caption{CP values (for the confidence ellipsoid with level $1-\xi=0.9$ for $\theta$) as a function of the training size $n$ of IS-AL and MLE when IM = PGA. 
    $R=100$ IS-AL replications are used to estimate the CP for each training size $n$. $B=200$ bootstrap samples are generated for the MLE to build the confidence ellipsoid at level $1-\xi=0.9$ for $n$ between $200$ and $500$, $B = 300$ bootstrap samples are generated for $n = 100$ due to numerical instabilities.}
    \label{fig:pic plot}
\end{figure}


Figure \ref{fig:pv plot} shows the CEVs for the MLE and IS-AL estimators. \cy{For $R = 200$ replications, these results show that for all the values of $n$ considered $\text{CEV}_{n}^{\rm IA} < \text{CEV}_{n}^{\rm MLE}$.} This indicates that MLE and IS-AL succeed in generating confidence ellipsoids that have the required coverage probability but MLE does so by generating ellipsoids that are much larger than the ones generated by IS-AL.
We can then conclude that IS-AL is much more efficient.

\begin{figure}[!ht]
    \centering
    \includegraphics[width=9cm]{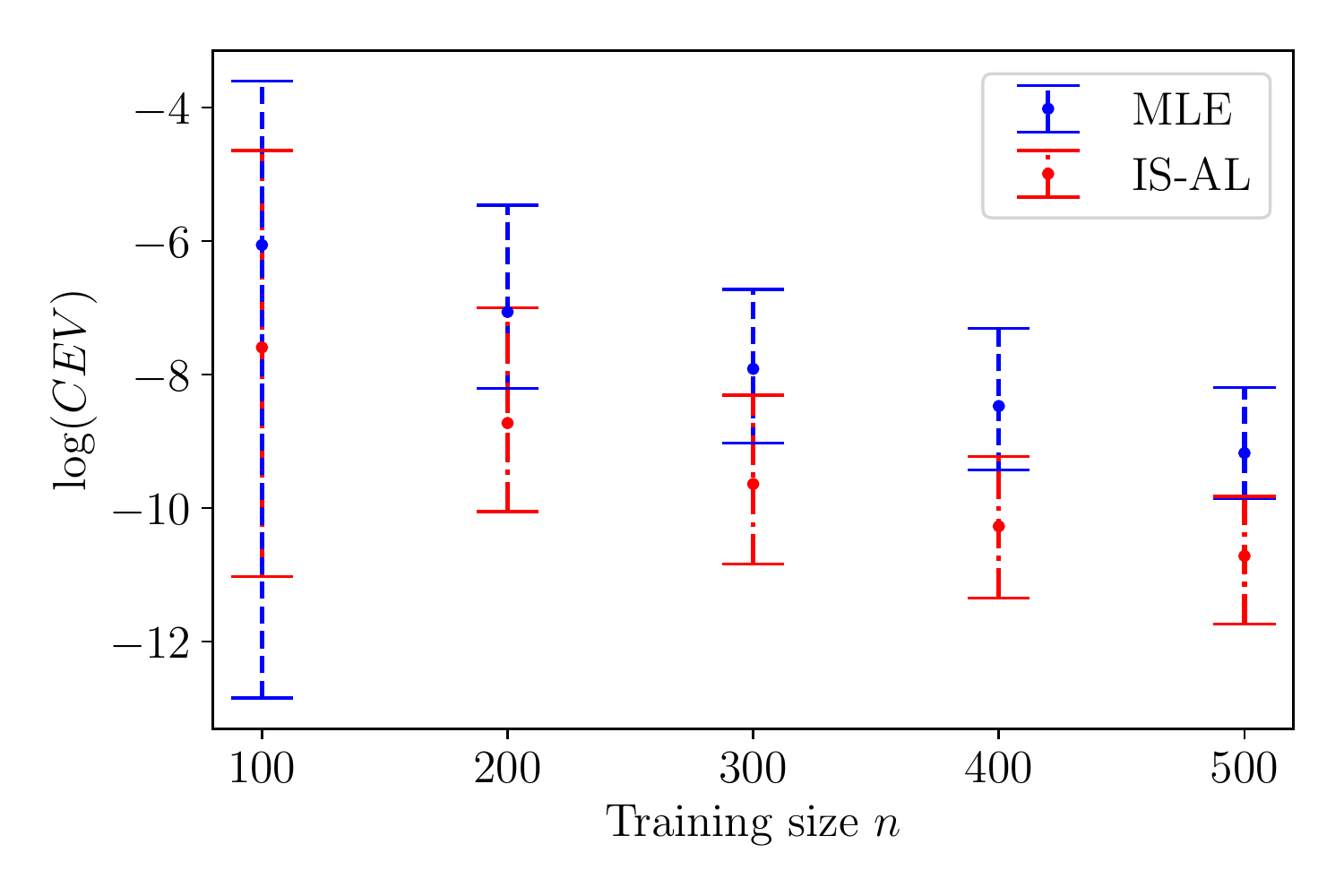}
    \caption{\modif{CEVs (for the confidence ellipsoid with level $1-\xi=0.9$ for $\theta$) as a function of the training size $n$ for IS-AL and MLE strategies when IM = PGA. The points are the medians over the $R=200$ replications while the vertical lines are the ranges between the $10\%$ and $90\%$ quantiles.}}
    \label{fig:pv plot}
\end{figure}
}

We emphasize that the convergence criterion $\widehat{W}_n$, illustrated in Figure~\ref{fig:stop_crit}, gives us at which sample size the IS-AL reaches asymptotic normality and thus at which sample size asymptotic confidence ellipsoid can be used. \cy{Even though $\text{CP}_{100}^{\rm IA}$ is less than the theoretical $90\%$, $70\%$ is considered as acceptable in practice.}

\subsubsection{Empirical distributions of the fragility curves and influence of the IM value \label{sec:empFC_inflIM}} 

\begin{figure}[!ht]
    \centering
    \includegraphics[width=9cm]{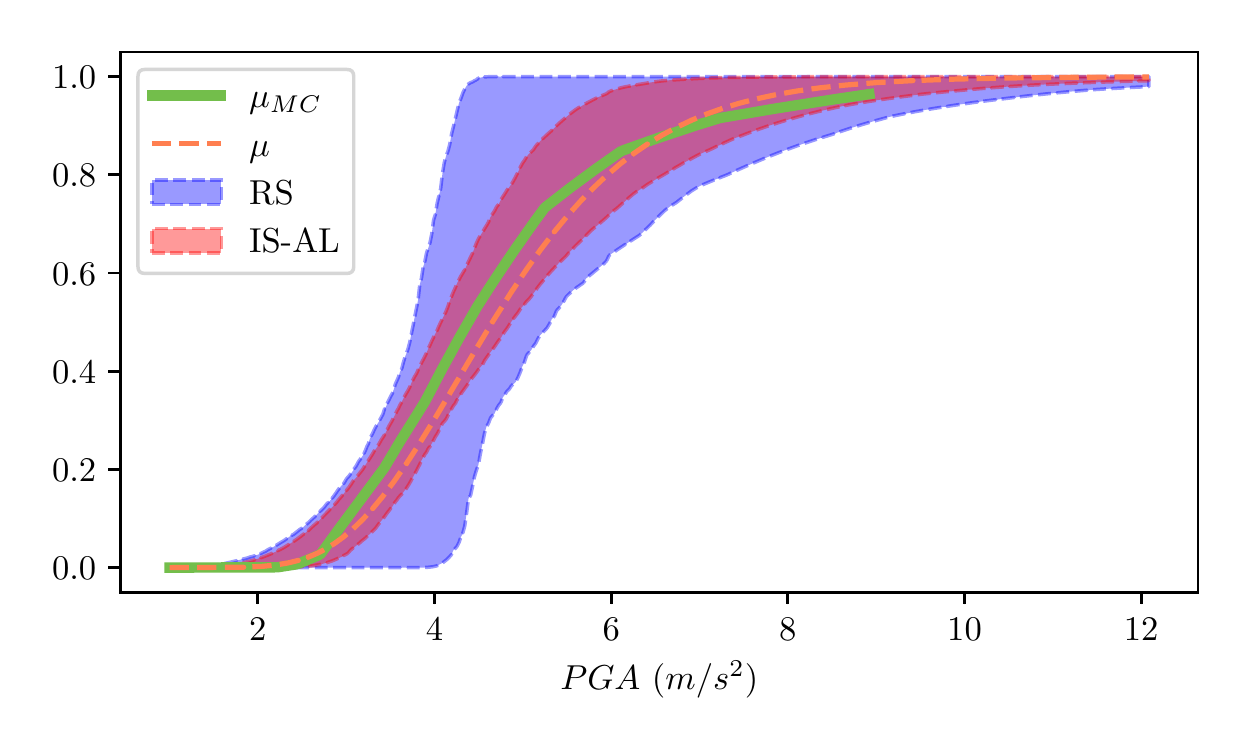}
    \caption{\modif{Empirical distribution of the fragility curves estimated by} RS and IS-AL for the nonlinear oscillator. The dashed orange line and solid green line are respectively the parametric estimation $\mu$ using $10^4$ seismic ground motions and the k-means nonparametric estimation of the fragility curve using $10^5$ seismic ground motions $\mu_{MC}$. The red and blue shaded areas correspond respectively to the $90\%$ to $10\%$ quantile ranges for the fragility curve dataset computed with IS-AL or \R{RS}. Remark that the nonparametric fragility curve is only plotted for \R{$PGA < 10 m/s^2$} due to the lack of seismic signals with PGA above that threshold.}
    \label{fig:kh_ci}
\end{figure}

\begin{figure}[!ht]
    \centering
    \includegraphics[width=9cm]{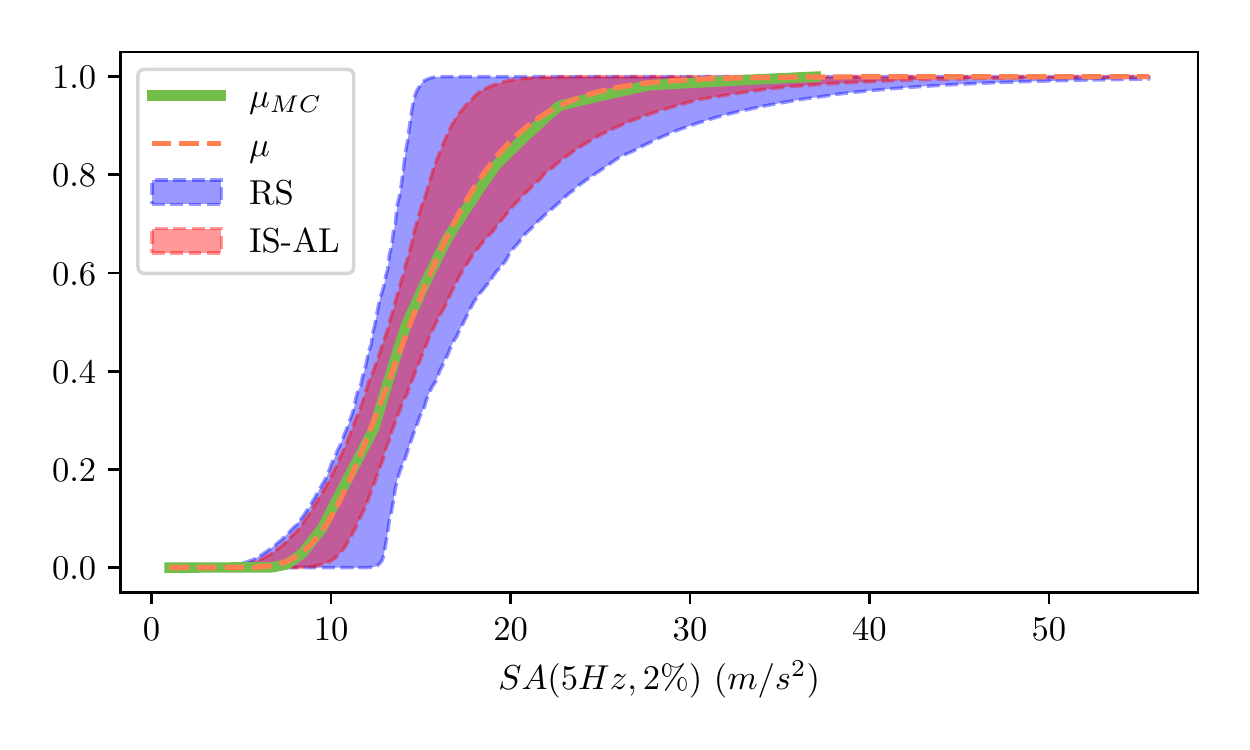}
    \caption{\cy{Empirical distributions of the fragility curves of RS and IS-AL for the nonlinear oscillator when IM is the spectral acceleration at $5$ Hz and $2\%$ damping ratio. The notations are the same as those in the Figure \ref{fig:kh_ci}. Remark that the bias between the nonparametric fragility curve $\mu_{MC}$ and the parametric fragility curve $\mu$ is smaller than the one obtained when using the PGA as the intensity measure (compare with Figure \ref{fig:kh_ci}).}}
    \label{fig:kh_ci_sa}
\end{figure}

\cy{The choice of the seismic IM is crucial for the accuracy of fragility curves estimates, especially when parametric models are concerned. So, empirical distributions of the fragility curves for IS-AL and RS methods are shown in Figure~\ref{fig:kh_ci} when IM = PGA, and in Figure~\ref{fig:kh_ci_sa} when IM is the spectral acceleration (SA) at $5$ Hz and $2\%$ damping ratio. The parametric fragility curves estimated with a dataset of $10^4$ seismic ground motions, are also shown in order to validate both the model choice and the uncertainty reduction provided by IS-AL.}

With the PGA, a bias between the lognormal fragility curve and the k-means nonparametric fragility curve, called $\mu_{MC}$, can be seen in Figure \ref{fig:kh_ci}. This phenomenon could be explained by the small correlation between maximal displacement of the oscillator during the seismic excitation and the PGA, which conveys small information about the seismic ground motion \cite{Ciano2020}.

 \R{The results  presented in Figure \ref{fig:kh_ci_sa} show a reduction of the bias between the nonparametric and the parametric fragility curve}. \R{This illustrates that, for the class of structures and for the seismic signal generator considered in this study, the parametric lognormal model has a better fit with the reference SA-based fragility curve than with reference PGA-based fragility curve.}

\modif{

\subsubsection{\cy{Confidence interval for parametric fragility curves : towards the engineering practice}\label{sec:isal_ci}}
\cy{After assessing the validity of the asymptotic confidence ellipsoid for IS-AL thanks to the computation of the CP values in section~\ref{sec:emp_CP_CEV}, we can use the asymptotic Gaussian distribution to construct the CI of the parametric fragility curve, as in the engineering practice. Thus, using a single run of the IS-AL procedure, we estimate the asymptotic covariance matrix $\widehat{G}_n$ and sample fragility curve parameters from the asymptotic distribution $\mathcal{N}\left(\widehat{\theta}_n^{\rm IA}, \frac{\widehat{G}_n}{n}\right)$. For the sake of comparison, we also construct the CI on a single replication of the MLE procedure using the bootstrap technique.

Figure \ref{fig:fragility kh CIs} represents the CIs for the fragility curve at level $90\%$ for IS-AL and MLE strategies, for a single replication of size $n=200$ of each procedure. Remark that the fragility curves estimated by MLE can be degenerated (i.e. as a unit step function), which implies that the CI for MLE is too conservative. This is consistent with the results of the figures~\ref{fig:kh mle params} and \ref{fig:pv plot}}.

\begin{figure}[!ht]
    \centering
    \includegraphics[width=9cm]{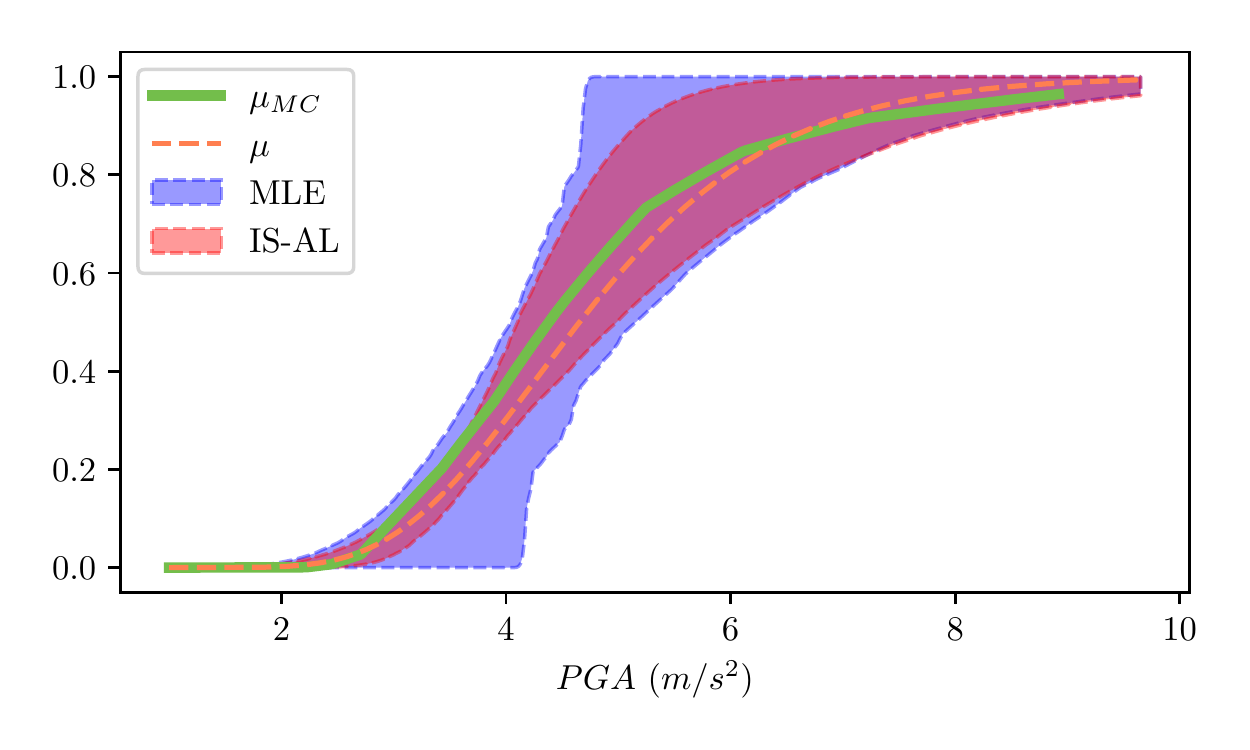}
    \caption{\modif{CIs of the parametric fragility curves of the elasto-plastic oscillator obtained with $500$ samples of the parameter asymptotic Gaussian distribution for IS-AL and $500$ bootstraped estimators with MLE. The red and blue shaded areas correspond to the ranges between the $95\%$ and $5\%$ quantiles for respectively IS-AL and MLE. The solid green line corresponds to the Monte Carlo estimation (k-means nonparametric estimation) of the fragility curve based on a dataset of $10^5$ of synthetic seismic signals.}}
    \label{fig:fragility kh CIs}
\end{figure}
}

\subsubsection{Synthesis}

\R{In this section, we have shown that the IS-AL-based methodology is (i) efficient to reduce the variance of the fragility curve estimation and (ii) can be applied regardless of the IM of interest. However, in practice, it is more suitable to use an IM as correlated as possible to the response of the structure to minimize potential biases due to the use of a parametric model. In addition, we have shown that, if the computation times allow it, it is possible to know when to stop the IS-AL algorithm, in order to build asymptotic confidence ellipsoids.}

\subsection{Industrial test case: safety water supply pipe of a pressurized water nuclear reactor}
\label{sec:industestcase}

\subsubsection{\cy{Description of the piping system}}

\begin{figure}[!ht]
		\centering		
		\subfloat[\label{fig:ASG_MU}]{\includegraphics[width=6cm]{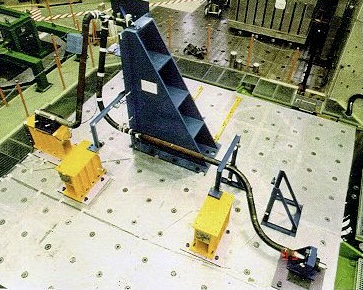}}
		\hspace{0.5cm}
		\subfloat[\label{fig:ASG_FEM}]{\includegraphics[trim= 1cm  3.8cm 12cm 1.5cm, clip,width=5cm]{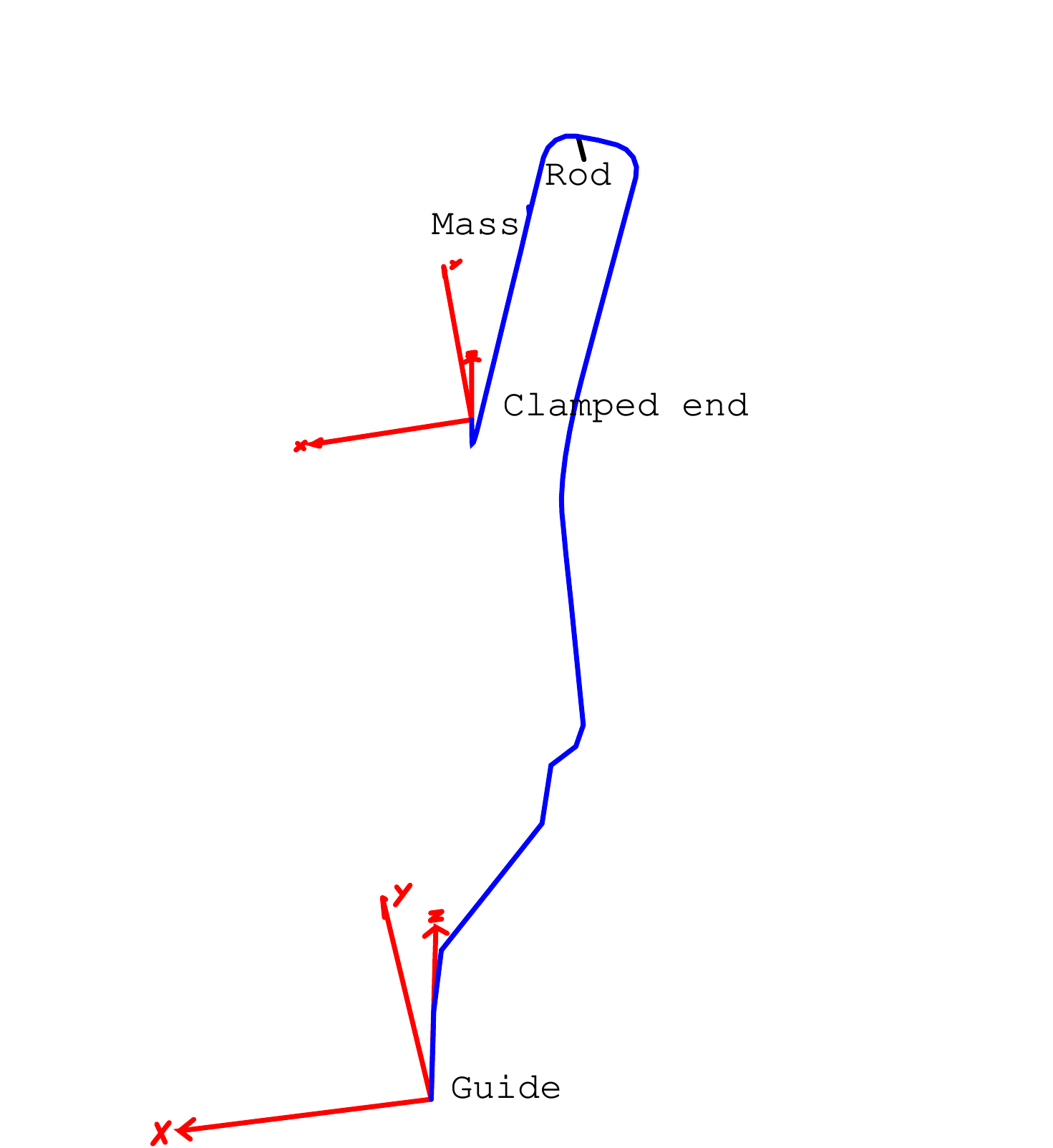}}
		\caption{(a) Overview of the ASG mock-up on the CEA's shaking table and (b) ASG FE model.}
		\label{fig:ASG}
	\end{figure}

The following test case corresponds to a piping system which is a simplified part of a secondary line of a French Pressurized Water Reactor. The numerical model was validated based on seismic tests performed on the shaking table Azalee of the EMSI laboratory of CEA/Saclay. The experimental program, called ASG program, and the main results are outlined in Ref.~\cite{TOUBOUL1999}. In Figure~\ref{fig:ASG_MU} a view of the mock-up mounted on the shaking table is shown. The Finite Element (FE) model, based on beam elements, is depicted in Figure~\ref{fig:ASG_FEM}.

The mock-up is a 114.3 mm outside diameter and 8.56 mm thickness pipe with a 0.47 elbow characteristic parameter, in carbon steel TU42C, filled with water without pressure. It contains three elbows and a mass modeling a valve (120 kg) which corresponds to more than 30\% of the specimen total mass. As shown in Figure~\ref{fig:ASG_FEM}, one end of the mock-up is clamped whereas the other is supported by a guide in order to prevent the displacements in the X and Y directions. Additionally, a rod is  placed on the top of the specimen in order to limit the mass displacements in the Z direction. In the tests, excitation act in the X direction.

Numerical comparisons are carried out with the homemade FE code CAST3M \cite{CAST3M}. Concerning the FE model, the boundary conditions are adjusted in order to obtain shapes and frequencies similar of those of the first two eigenmodes of the mock-up in the X and Y directions, respectively at 5.1 Hz and 6.6 Hz. As measured in the experiments, a critical damping ratio of 1\% is considered for these two eigenmodes with a damping Rayleigh assumption. Finally, regarding the nonlinear constitutive law of the material, a bilinear law exhibiting kinematic hardening is used to reproduce the overall nonlinear behavior of the mock-up with satisfactory agreement compared to the results of seismic tests  \cite{TOUBOUL1999}. 

In the context of this test case, the yield stress of the bilinear law is equal to $3 \; 10^8$~Pa, the Young modulus is equal to $1.92 \; 10^{11}$~Pa whereas the hardening modulus is equal to $4.3 \; 10^8$~Pa. Moreover, since for the synthetic signals considered in this work (the same as those used in the reference \cite{Sainct20} and in the second test case of this paper) the piping system remains in the linear domain, they are filtered by a fictitious linear single-mode building at $5$ Hz and damped at $2\%$. Finally, we consider excessive out-of-plane rotation of the elbow located near the clamped end of the mock-up as failure criterion, as recommended in \cite{TOUBOUL2006}. Since the weight of the mass is not completely taken up by the mechanical assembly, the overall behavior of the mock-up exhibits ratcheting.

In the following, the random variable $R_e$ corresponds to the maximum of the out-of-plane rotation of the elbow. The binary variable which indicates the failure state is defined by $S = {\mathbf 1}_{R_e>C}$ where $C$ is the admissible rotation in degree. In our case, $C = 4.38^{\circ}$. This value is the $90\%$-level quantile from a sample of $2000$ mechanical simulations. 

\subsubsection{\cy{Performance metrics}}

\begin{figure}[!ht]
     \centering
     \begin{subfigure}[b]{0.48\textwidth}
         \centering
         \includegraphics[width=\textwidth]{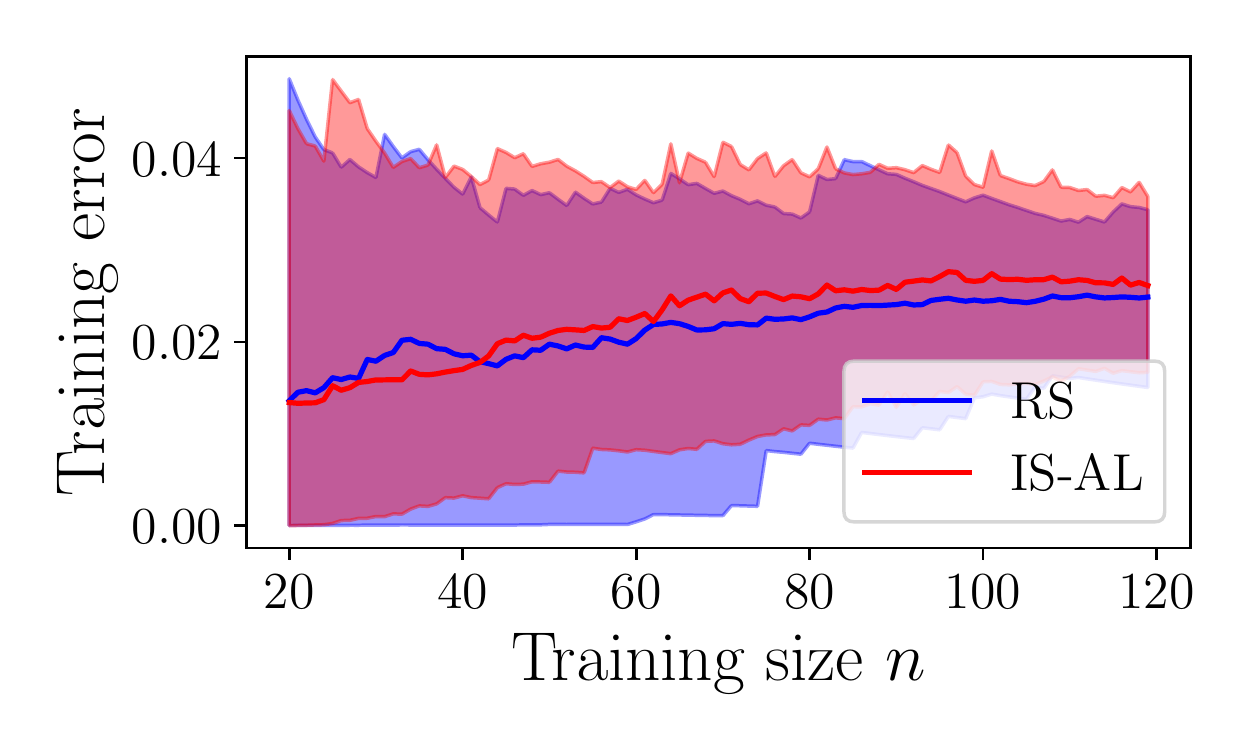}
         \caption{$\widehat{R}^{\rm IA}_{n}$ versus $\widehat{R}^{\rm RS}_{n}$}    
         \label{fig:asg train}
     \end{subfigure}
     \hfill
     \begin{subfigure}[b]{0.48\textwidth}
         \centering
         \includegraphics[width=\textwidth]{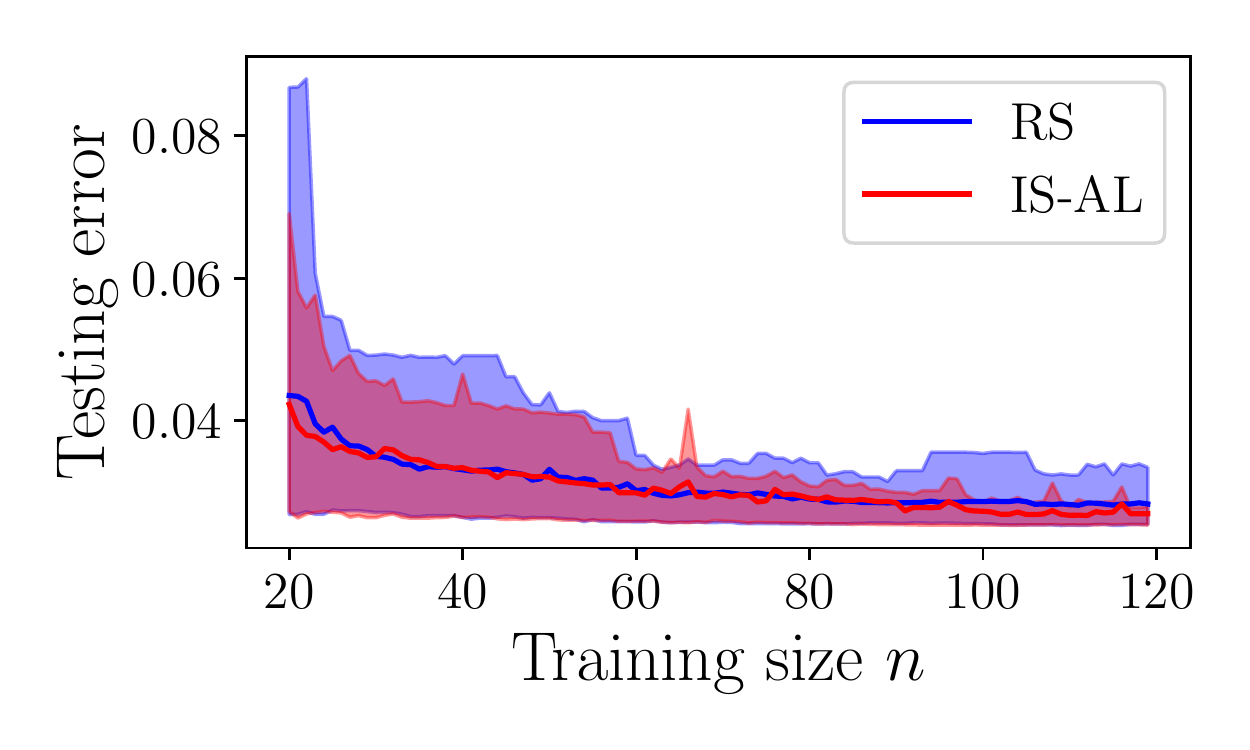}
         \caption{$\widehat{Q}^{\rm IA}_{n}$ versus $\widehat{Q}^{\rm RS}_{n}$}                  
         \label{fig:asg test}
     \end{subfigure}
     \begin{subfigure}[b]{0.48\textwidth}
         \centering
         \includegraphics[width=\textwidth]{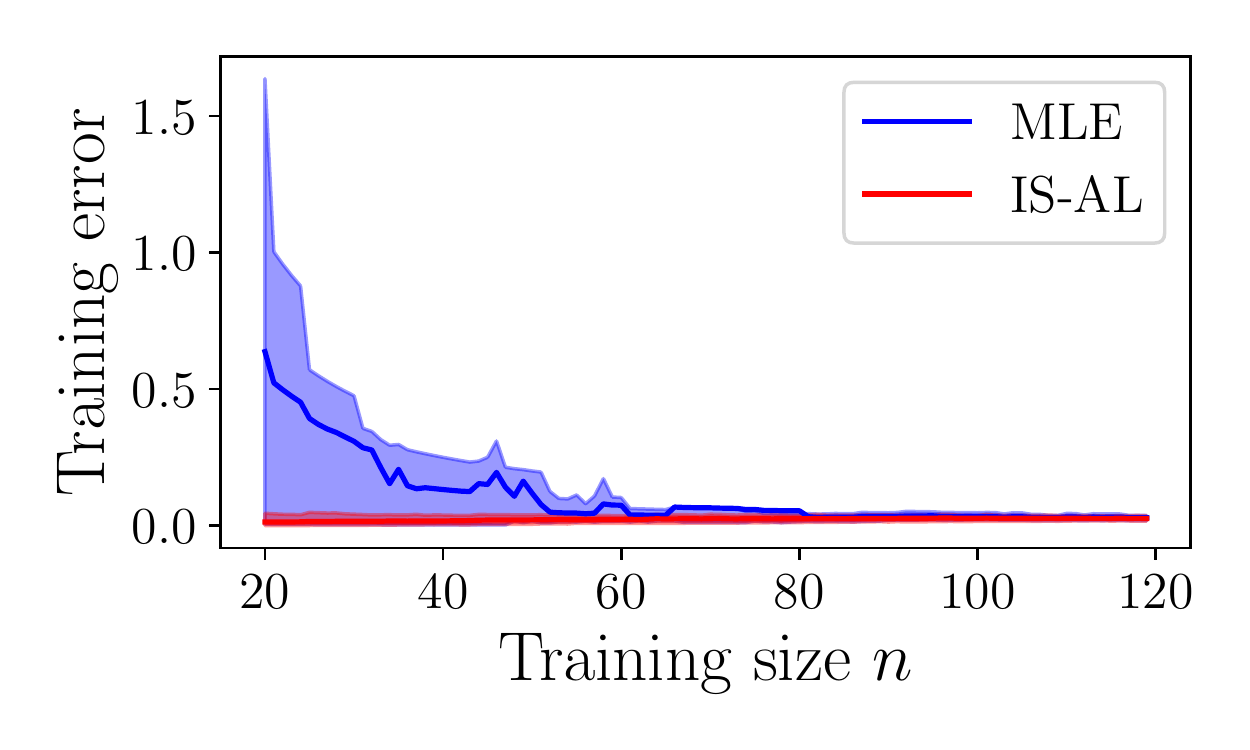}
         \caption{$\widehat{R}^{\rm IA}_{n}$ versus $\widehat{R}^{\rm MLE}_{n}$}          
         \label{fig:asg mle train}
     \end{subfigure}
     \hfill
     \begin{subfigure}[b]{0.48\textwidth}
         \centering
         \includegraphics[width=\textwidth]{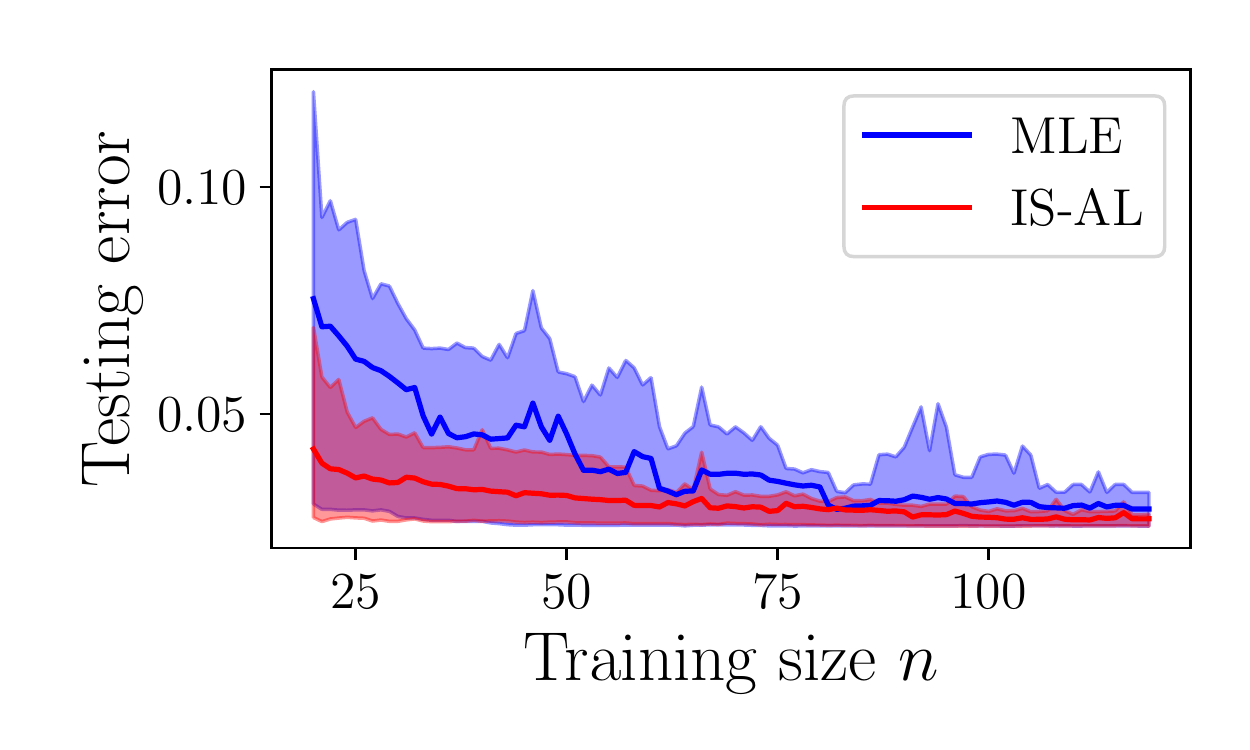}
         \caption{$\widehat{Q}^{\rm IA}_{n}$ versus $\widehat{Q}^{\rm MLE}_{n}$}          
         \label{fig:asg mle test}
     \end{subfigure}     
     \caption{Numerical benchmark of the ASG piping system. \modif{The empirical distributions of the training and testing errors are computed, the red and blue shaded areas correspond to the area between the empirical quantiles of levels $10\%$ and $90\%$ of the $50$ replications for respectively IS-AL, RS and MLE}.}
     \label{fig:asg loss}
\end{figure}

\cy{For this test case, the numerical benchmark is based on $50$ replications of $120$ signals sampled using IS-AL (that includes the initial $20$ points) with a defensive parameter $\varepsilon=10^{-3}$ and $120$ signals for the RS and MLE strategies.

The IS-AL procedure is initialized by considering the linear FE model of the ASG mock-up and a numerical resolution based on a modal base projection. Thus, the initialization parameter $\widehat{\theta}_0^{\rm IA}$ is approximated by $\widehat{\theta}^{{\rm RS}}_{2000}$ using a $2000$-sized dataset randomly selected from the $10^5$ synthetic seismic ground motions. Then, $20$ datapoints are queried with the instrumental density $q_{{\widehat{\theta}}^{\rm IA}_0,\varepsilon}$ before launching the adaptive strategy. For the training, $100$ signals are then chosen in a pool of $1500$ CAST3M computations while $500$ additional computations are carried out in order to compute the testing error.

Figure \ref{fig:asg loss} compares the IS-AL, MLE and RS training and testing errors as functions of $n$. Remark that the training loss of MLE is greater than the training loss of IS-AL. This numerical artifact is essentially due to the regularization term $\beta_{\rm reg}$: if the $\beta$ parameter estimated by MLE is small, the penalization term ${\beta_{\rm reg}}/{\beta}$ can be very high.

Table \ref{table:asg} shows that the IS-AL strategy has overall better performance than the other two strategies.}

\begin{table}[!ht]
    \caption{Performance metrics for the ASG piping system for $n = 120$ when IM = SA}
    \label{table:asg}
    \cy{\begin{center}
        \begin{tabular}{ c|c|c|c||c|c|c } 
         \multicolumn{1}{c}{} & \multicolumn{3}{c}{Train} & \multicolumn{3}{c}{Test} \\
         \hline
         {${\rm \bullet}$} & RS & MLE & {IS-AL} & RS & MLE & {IS-AL} \\ 
         \hline
          {$\text{RSD}_{120}^{\rm \bullet}$} ($\%$) & $40.5$ & $46$ & $34$ & $24.1$ & $28$ & $12$ \\
          {$\nu_{120}^{\rm \bullet}$} & $0.93$ & $2.4$ & $\times$ & $1.3$ & $5.8$ & $\times$ \\
          {$\text{RB}_{120}^{\rm \bullet}$} ($\%$) & $7.2$ & $8.6$ & $5.5$ & $18$ & $8.4$ & $0.3$
        \end{tabular}
    \end{center}}
\end{table}

\subsubsection{Empirical distributions of the parameters $\alpha$ and $\beta$}

\begin{figure}[!ht]
    \centering
    \includegraphics[width=9cm]{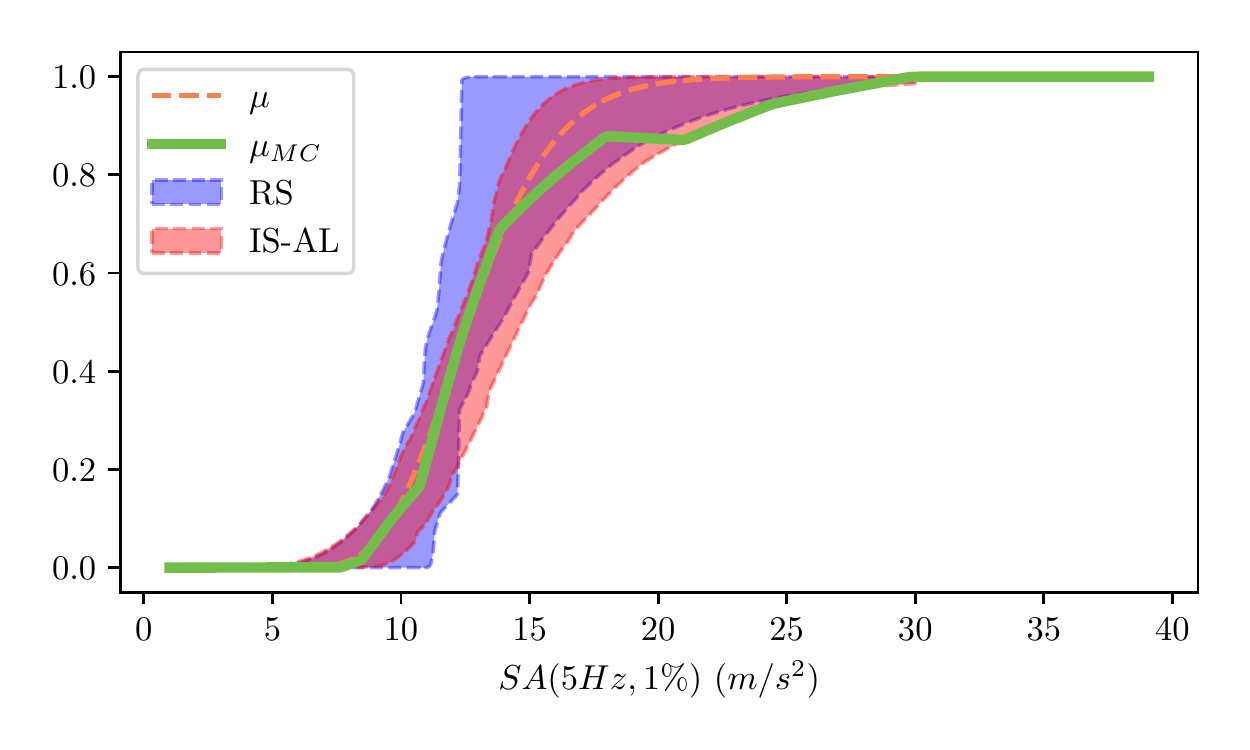}
    \caption{\modif{Empirical distributions of the fragility curves of RS and IS-AL for the ASG piping system}. The red and blue areas correspond respectively to the ranges between the $10\%$ and \R{$90\%$} quantiles of the fragility curve dataset generated with IS-AL and RS with $n=120$ training datapoints (that includes the initialization points). The dashed orange line corresponds to a parametric fragility curve estimation using least squares minimization with a dataset of $2000$ seismic ground motions and FE simulations of the piping system. \modif{The solid green line corresponds to a Monte Carlo estimation (k-means nonparametric estimation) of the fragility curve using the same $2000$-sized dataset.}}
    \label{fig:asg_ci}
\end{figure}

 \modif{Figure \ref{fig:asg mle params} compares the distributions of parameters $\alpha$ and $\beta$ for several sample sizes between MLE and IS-AL using $50$ replications. As with the nonlinear oscillator, the $\beta$ parameter estimated with IS-AL is less likely to be close to $0$ than when it is estimated with MLE. This motivates further the use of active learning to have a better accuracy for fragility curves parameters estimates with the same computational cost as state of the art estimation methods.

\begin{figure}[!ht]
     \centering
     \begin{subfigure}[b]{0.48\textwidth}
         \centering
         \includegraphics[width=\textwidth]{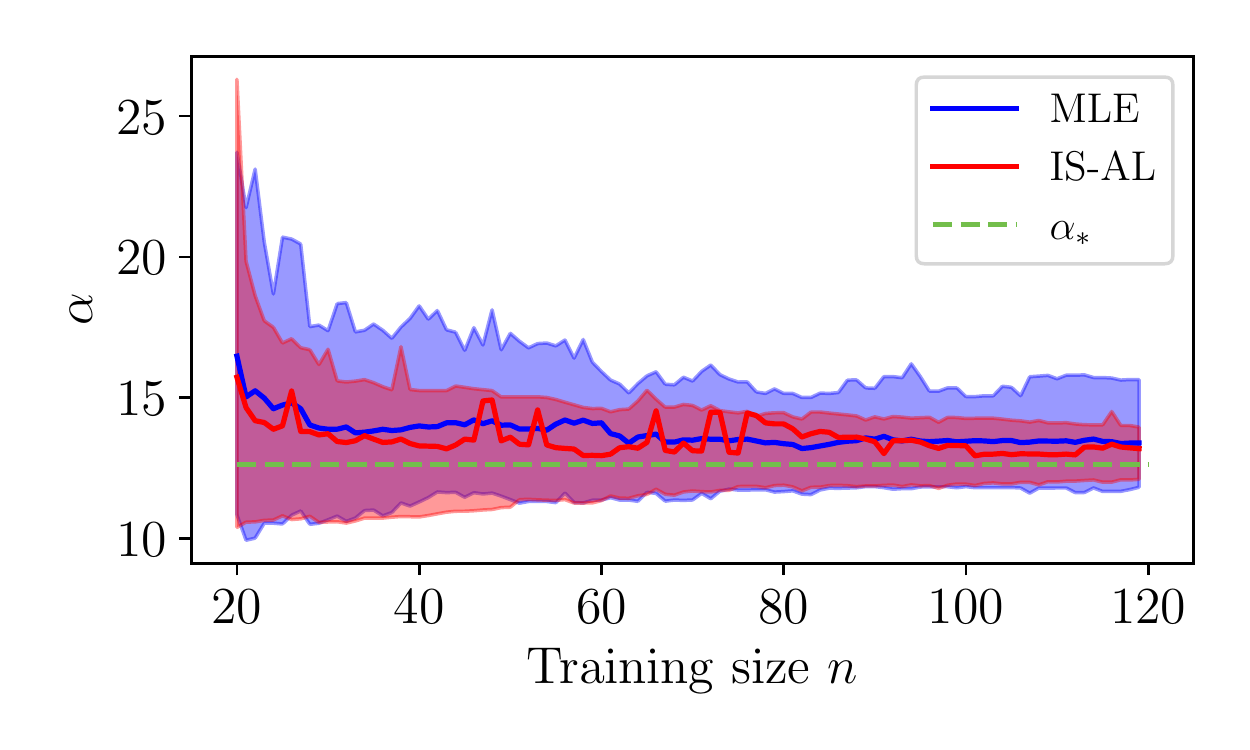}
     \end{subfigure}
     \hfill
     \begin{subfigure}[b]{0.48\textwidth}
         \centering
         \includegraphics[width=\textwidth]{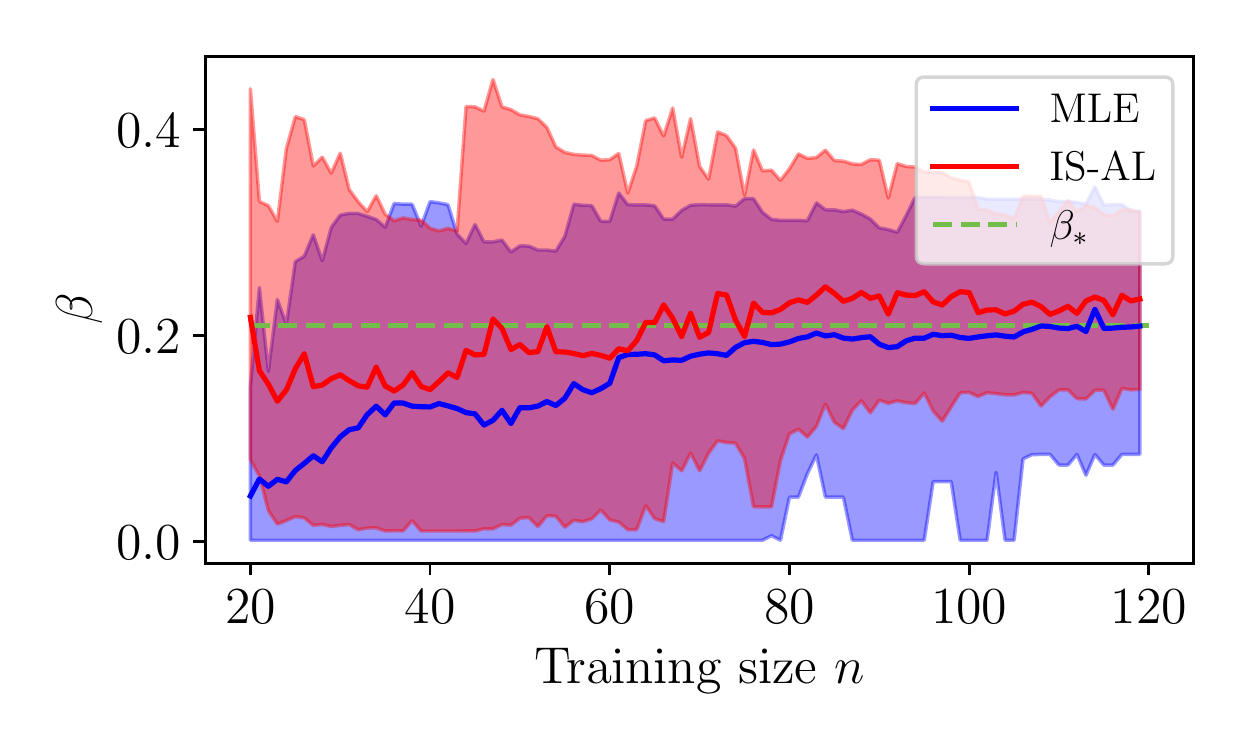}
     \end{subfigure}
     \caption{\modif{Results for the ASG piping system: the empirical distributions of parameters $\alpha$ and $\beta$ are represented by the empirical $90\%$ and $10\%$ quantiles of $50$ replications and correspond to the shaded blue and red areas respectively for MLE and IS-AL. The dashed green lines correspond to the values $\alpha_*$ and $\beta_*$, which have been here approximated by $\widehat{\alpha}_N, \ \widehat{\beta}_N$ for $N = 2000$.}}
     \label{fig:asg mle params}
\end{figure}

\subsubsection{Fragility curve estimations}

Figure \ref{fig:asg_ci} illustrates the uncertainty reduction provided by IS-AL on the fragility curve estimate. Motivated by the results obtained for the nonlinear oscillator, the fragility curve of the piping system is here expressed as a function of the pseudo-spectral acceleration of the initial set of synthetic signals (i.e not filtered signals), calculated at $5$ Hz and $1\%$ damping ratio.}

\begin{figure}[!ht]
    \centering
    \includegraphics[width=9cm]{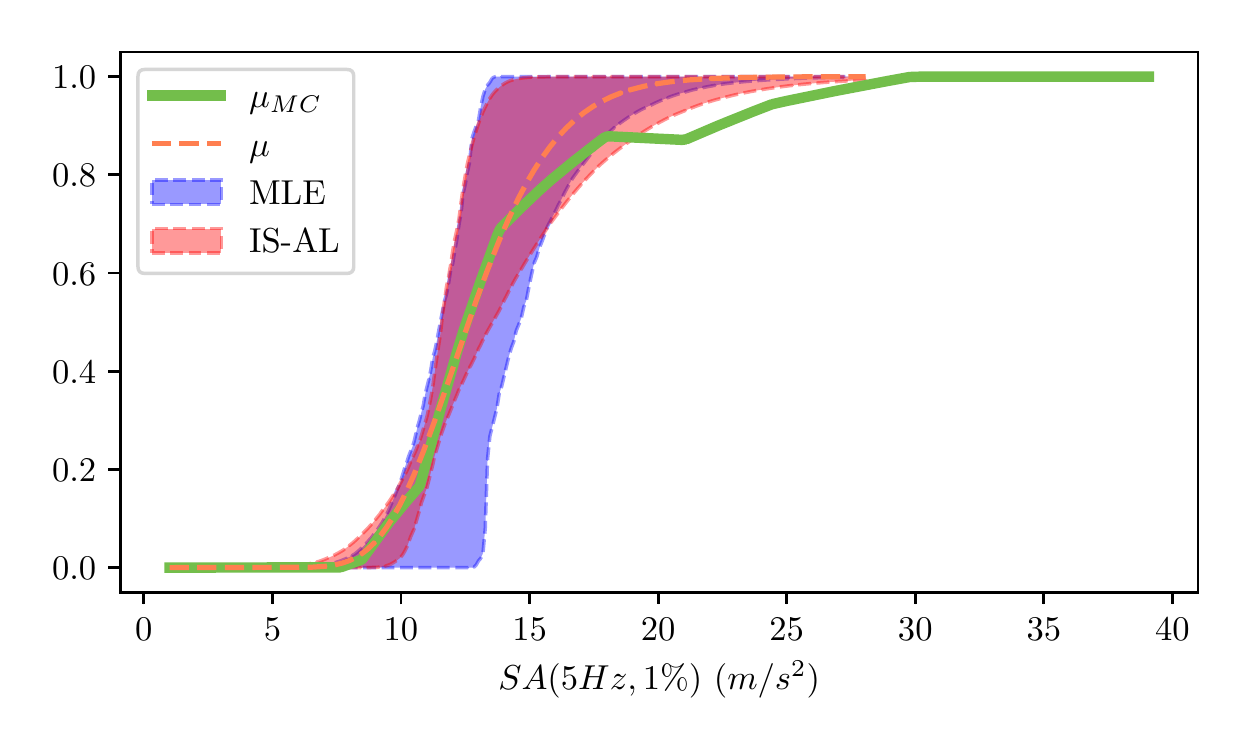}
    \caption{\modif{Parametric fragility curve of the ASG piping system confidence interval obtained with $500$ samples of the parameter asymptotic Gaussian distribution for IS-AL and $500$ bootstraped estimators with MLE. The red and blue shaded areas correspond to the ranges between the $95\%$ and $5\%$ quantiles for respectively IS-AL and MLE. The solid green line corresponds to a Monte Carlo estimation (k-means nonparametric estimation) of the fragility curve using our $2000$ sized dataset of CAST3M computations.}}
    \label{fig:asg_isal_ci}
\end{figure}

Figure \ref{fig:asg_isal_ci} represents the confidence interval on the fragility curve for IS-AL and MLE using a single replication of $200$ CAST3M computations ($20$ computations used for initialization and $180$ computations for IS-AL), obtained with the same methodology as that presented in section~\ref{sec:isal_ci}. Remark that the Monte Carlo estimation of the fragility curve (i) belongs to the confidence interval of IS-AL for seisms with relatively small spectral acceleration (ii) is not accurate for high spectral accelerations due to the lack of seismic signals of such intensities.

\cy{As for the nonlinear oscillator, the figures \ref{fig:asg_ci} and \ref{fig:asg_isal_ci} suggest that for the RS and MLE strategies, even with $n = 120$ points, it is possible to obtain samples for which a $\beta$ estimate is close to $0$, which IS-AL avoids.}

\section{Conclusion}
In this paper, we have introduced an original methodology to improve the accuracy of parametric fragility curve estimation without increasing the sample size, thanks to an active learning strategy \modif{based on importance sampling}. Defensive strategy has been implemented to control the likelihood ratio and the possible increase of the training loss variance in the early steps. We use a penalized least square loss to avoid an identifiability issue of the standard deviation of the lognormal model. 
We define a convergence criterion that indicates asymptotic normality of the estimator and provide asymptotic confidence intervals and ellipsoids.
We illustrate our active learning procedure in numerical examples, from a synthetic case to a FE mechanical simulation of a piping system of a French Pressurized Water Reactor.
\cy{In comparison with the engineering practice based on the joint use of the MLE and the boostrap techniques, the proposed methodology is more efficient. For the same number of calculations, the IS-AL procedure reduces the variance of the parametric estimation of the fragility curve and gives theoretical guarantees on the convergence of the estimations.}

\section*{Acknowledgements}

This research was supported by CEA (French Alternative Energies and Atomic Energy Commission) and SEISM Institute (www.institut-seism.fr/en/).\\

\appendix

\section{Proof of \modif{Equation \eqref{eq: consistency}}}
\label{app: consistency}
\modif{Throughout the appendix, the IS-AL estimator $\widehat{\theta}^{\rm IA}_n$ is denoted by $\tilde{\theta}_n$ and the loss $\widehat{R}^{\rm IA}$ by $\tilde{R}$.}

\modif{The proof for the consistency is based on Theorem 2 of \cite{Deylon18}. We precise the needed assumptions in a very general way, with a parametric family $\mathcal{F} = \{f_{\theta}, \ \theta \in \Theta \}$, loss function $\ell_{\theta}$ and instrumental density $q_{\theta}$. We will then check that the needed assumptions are satisfied by IS-AL. Set $L(x, s) = \sup_{\theta \in \Theta}\ell_{\theta}(x, s)$. Assume that $\Theta$ is a compact set, $\theta_{*} = \argmin_{\theta \in \Theta} r(\theta)$ exists and is unique and that:
$$
\iint L(x, s) P(dx,ds) < +\infty ,
$$
$$
\sup_{\theta \in \Theta} \iint \frac{L(x, s)^2p(x)}{q_{\theta}(x)}P(dx,ds) < +\infty ,
$$
$$
\forall \theta \neq  \theta_{*}, \iint \ell_{\theta}(x, s)P(dx, ds) > \iint \ell_{\theta_{*}}(x, s)P(dx, ds) .
$$
and for any $(x, s) \in {\cal X} \times \{0, 1\}$, $\theta \in \Theta \mapsto \ell_{\theta}(x, s)$ is continuous. Thus, we can apply Theorem 2 of \cite{Deylon18} in order to prove the consistency of $\tilde{\theta}_n = \argmin_{\theta \in \Theta} \frac{1}{n} \sum\limits_{i=1}^n \frac{p(X_i)}{q_{\tilde{\theta}_{i-1}}(X_i)} \ell_{\theta}(X_i, S_i)$. More precisely, these assumptions are verified for IS-AL. Indeed, the regularized squared loss is bounded for the variables $\theta, \ x, s$ when $\theta = (\alpha, \beta)^T$ is in a compact set of $(0,+\infty)^2$. Moreover, the likelihood ratio $\frac{p(x)}{q_{\theta, \varepsilon}(x)}$ with the defensive instrumental density is bounded for $x \in \mathcal{X}$. Concerning the regularization, we have:
\begin{equation}
    \left|\tilde{R}_n(\theta) - r(\theta)\right| < \left|\frac{1}{n} \sum\limits_{i=1}^n \frac{p(X_i)}{q_{\modif{\widehat{\theta}_{i-1}}^{\rm IA}, \varepsilon}(X_i)} \ell_{\theta}(X_i, S_i) - r(\theta)\right| + \frac{\Omega(\theta; \beta_{\rm reg})}{n} \ .
\end{equation}
Thus the condition $(17)$ of Theorem 2 in \cite{Deylon18} is still valid.
}

\section{Proof of \modif{the asymptotic normality of $\modif{\widehat{\theta}_n^{\rm IA}}$}}
\label{app: normality}
\modif{In the same way as in the proof of the consistency of $\tilde{\theta}_n$, we provide a general proof of asymptotic normality. Assume that $\theta \mapsto \ell_{\theta}$ is three times differentiable at $\theta_*$ for all $x, s$ and that the matrix $\ddot{r}(\theta_{*})$ exists and is nonsingular. Assume that the third-order derivatives of $\theta\mapsto \ell_{\theta}(x, s)$ are dominated in a neighborhood of $\theta_{*}$ by a function that is integrable with respect to $P$.
Assume also that the following conditions are satisfied:
\begin{enumerate}
    \item  The hypotheses needed for the consistency of $\tilde{\theta}_n$ are satisfied,
    \item  $\exists \eta > 0 $ such that  $\sup_{\theta \in \Theta}\iint ||\frac{p(x)\dot{\ell}_{\theta_{*}}(x, s)}{q_{\theta}(x)}||^{2+\eta} P(dx,ds) < +\infty$,
    \item  $ \sup_{\theta \in \Theta}\iint \frac{p(x) || \ddot{\ell}_{\theta_{*}}(x,s)\ddot{\ell}_{\theta_{*}}(x,s)^T||}{q_{\theta}(x)} P(dx,ds) < +\infty$,
    \item  there exists a neighborhood $\mathcal{B}$ of $\theta_{*}$ such that $\forall (x,s) \in \mathcal{X}\times\{0,1\}$,\\ $\sup_{\theta \in \mathcal{B}}\frac{p(x)\|\dddot{\ell}_{\theta}(x,s)\|}{q_{\theta}(x)} < +\infty$.
\end{enumerate}}
The asymptotic normality of an estimator built such as $\tilde{\theta}_n$ is based on the following arguments highlighted in Theorem 5.41 of \cite{vaart1998}:
\begin{itemize}
    \item \textbf{(P1)} The random function $\sqrt{n}\Psi_n(\theta_*)$, with $\Psi_n(\theta) = \dot{\tilde{R}}_n(\theta) - \dot{r}(\theta) $, converges in law to a centered Gaussian distribution with covariance $V_{\theta_*}$.
    \item \textbf{(P2)} The random function $\dot{\Psi}_n(\theta_*)$ converges in probability to  $\mathbb{E}[\ddot{\ell}_{\theta_*}(X,S)]$ 
    \item \textbf{(P3)} The random function $\ddot{\Psi}_n(\theta_n)$ is bounded in probability for $\theta_n$ a deterministic sequence in a neighborhood of $\theta_*$.
\end{itemize}
Of course, we need all the quantities above to be properly defined, hence we have to restrict ourselves to a loss function $\theta \mapsto \ell_{\theta}$ that is smooth enough, such as the quadratic loss. \modif{We use Theorem 1 of \cite{Deylon18} to prove proposition $\textbf{(P1)}$.} Theorem 2.18 in \cite{hall2014martingale} ensures that \textbf{(P2)} and \textbf{(P3)} are verified by the assumptions $2)$, $3)$ and $4)$ so that $\dot{\Psi}_n(\theta_*)$ converges toward the matrix $\ddot{r}(\theta_*)$. The sequence $\sqrt{n}(\tilde{\theta}_n-\theta_*)$ is asymptotically normal with mean zero and covariance matrix $\ddot{r}(\theta_*)^{-1}V_{\theta_*} (\ddot{r}(\theta_*)^{-1})^{T}$. \modif{For IS-AL, the functions $\dot{\ell}_{\theta}$, $\ddot{\ell}_{\theta}$, $\dddot{\ell}_{\theta}$ are continuous for variables $\theta, \ x$ on a compact set and thus are bounded for variable $\theta$, in the same way as for the consistency, the likelihood ratio for the defensive instrumental density is bounded for $x \in \mathcal{X}$. Concerning the regularization, the third derivative $\dddot{\Omega}(\theta; \beta_{\rm reg})$ is continuous on $\Theta$ which is compact, hence bounded. Naturally, we have \textbf{(P3)} verified. Because $\frac{\dot{\Omega}(\theta; \beta_{\rm reg})}{n}$ converges in probability to $0$, \textbf{(P2)} is also verified. Using Slutsky's lemma, \textbf{(P1)} is verified.}

\section{Proof of Lemma \modif{Equation \eqref{eq: Gn consistency}}}
\label{app:prooflemmaGn}
\modif{First of all, we precise the needed assumptions for a general proof. Set $L_{1,k,l}(x,s) = sup_{\theta \in \Theta} \ddot{\ell}_{\theta}(x, s)_{k, l}$ and $L_{2,k,l}(x, s) = sup_{\theta \in \Theta} \frac{p(x)}{q_{\theta}(x)} (\dot{\ell}_{\theta}(x,s) \dot{\ell}_{\theta}(x,s)^T)_{k, l}$ $\forall k,l = 1,...,m$ and assume that:
\begin{enumerate}
    \item $\inf_{(\theta, x, s ) \in \Theta \times \mathcal{X} \times \{0, 1\}} \frac{p(x)}{q_{\theta}(x)} \ddot{\ell}_{\theta}(x, s)_{k,l}  > -\infty$ $\forall k,l=1,\ldots,m$.
    \item $\inf_{(\theta, x, s ) \in \Theta \times \mathcal{X} \times \{0, 1\}} \left(\frac{p(x)}{q_{\theta}(x)}\right)^2 \dot{\ell_{\theta}}(x,s)\dot{\ell_{\theta}}(x,s)^T_{k,l}  > -\infty$ $\forall k,l=1,\ldots,m$.
    \item $\iint L_{i,k,l}(x, s) P(dx,ds) < +\infty$, $\forall i \in \{1, 2\}$, $\forall k,l = 1,...,m$.
    \item $\sup_{\theta \in \Theta} \iint \frac{L_i(x, s)^2p(x)}{q_{\theta}(x)}P(dx,ds) < +\infty, \  i \in \{1, 2\}$.
\end{enumerate}}
The result comes from the uniform convergence of $\widehat{G}_n(\theta) = \widehat{\ddot{r}}_n(\theta)^{-1} \widehat{V}_n(\theta)(\widehat{\ddot{r}}_n(\theta)^{-1})^T$ to $G_{\theta}$ for $\theta$ in a neighborhood of $\theta_*$. It boils down to prove uniform convergence of $\widehat{\ddot{r}}_n(\theta)$ and $\widehat{V}_n(\theta)$. The proof is in the same spirit as in \modif{\ref{app: consistency}}. We proceed coordinate by coordinate defining $H_i(\theta)_{k, l} = \frac{p(X_i)}{q_{\theta}(X_i)} \ddot{\ell}_{\theta}(X_i, S_i)_{k,l} - \inf_{(\theta, x, s ) \in \Theta \times \mathcal{X} \times \{0, 1\}} \frac{p(x)}{q_{\theta}(x)} \ddot{\ell}_{\theta}(x, s)_{k,l}$ to prove uniform convergence of $\widehat{\ddot{r}}_n(\theta)$ and $H_i(\theta)_{k, l} = \left(\frac{p(X_i)}{q_{\theta}(X_i)}\right)^2 \dot{\ell}_{\theta}(X_i, S_i)\dot{\ell}_{\theta}(X_i, S_i)^T_{k,l} - \inf_{(\theta, x, s ) \in \Theta \times \mathcal{X} \times \{0, 1\}} \left(\frac{p(x)}{q_{\theta}(x)}\right)^2 \dot{\ell}_{\theta}(x,s)\dot{\ell}_{\theta}(x,s)^T_{k,l}$ for $\widehat{V}_n(\theta)$. Assumptions $3)$ and $4)$ ensure the uniform convergence using the proof technique of Theorem \modif{1 of \cite{Deylon18}}.

\section{Proof of \modif{Equation \eqref{eq: asymp grad}}}

The proof relies on the Taylor expansions of $\dot{\tilde{R}}_{n,1}(\tilde{\theta}_{n,2})$ and $\dot{\tilde{R}}_{n,2}(\tilde{\theta}_{n,1})$ around the parameter value $\theta_*$:
$$
\dot{\tilde{R}}_{n,1}(\tilde{\theta}_{n,2}) = \dot{\tilde{R}}_{n,1}(\theta_*) + \ddot{\tilde{R}}_{n,1}(\theta_*)(\tilde{\theta}_{n,2} - \theta_*) + o\left(\lVert\tilde{\theta}_{n,2} - \theta_*\rVert\right) . 
$$
$$
\dot{\tilde{R}}_{n,2}(\tilde{\theta}_{n,1}) = \dot{\tilde{R}}_{n,2}(\theta_*) + \ddot{\tilde{R}}_{n,2}(\theta_*)(\tilde{\theta}_{n,1} - \theta_*) + o\left(\lVert\tilde{\theta}_{n,1} - \theta_*\rVert\right) . 
$$
Using \modif{the asymptotic normality of $\tilde{\theta}_n$}, we can apply \modif{Appendix B.3 in \cite{Deylon18}} to prove the convergence of $\ddot{\tilde{R}}_{n,1}(\theta_*)$ and $\ddot{\tilde{R}}_{n,2}(\theta_*)$ to $\ddot{r}(\theta_*)$ in the same spirit as for the proof of \modif{\ref{app: consistency}}. We proceed coordinate by coordinate, defining $H_i(\theta_*)_{k,l} = \frac{p(X_i)}{q_{\tilde{\theta}_{i-1}(X_i)}}\ddot{\ell}_{\theta_*}(X_i, S_i)_{k,l} - \inf_{(\theta, x, s) \in \Theta \times \mathcal{X} \times \{0, 1\}} \frac{p(x)}{q_{\theta}(x)} \ddot{\ell}_{\theta_*}(x, s)_{k,l} \ .$ Remark that $H_i(\theta_*)_{k,l} \geq 0$, hence we can apply \modif{Appendix B.3 in \cite{Deylon18}} to obtain the desired convergence.  
Moreover, the Taylor expansions of $\dot{\tilde{R}}_{n,1}(\tilde{\theta}_{n,1})$ and $\dot{\tilde{R}}_{n,2}(\tilde{\theta}_{n,2})$ write:
\begin{align*}
0 &= \dot{\tilde{R}}_{n,1}(\tilde{\theta}_{n,1}) =  \dot{\tilde{R}}_{n,1}(\theta_*) + \ddot{r}(\theta_*)(\tilde{\theta}_{n,1} - \theta_*) + o\left(\lVert\tilde{\theta}_{n,1} - \theta_*\rVert\right),\\
0 &= \dot{\tilde{R}}_{n,2}(\tilde{\theta}_{n,2}) =  \dot{\tilde{R}}_{n,2}(\theta_*) + \ddot{r}(\theta_*)(\tilde{\theta}_{n,2} - \theta_*) + o\left(\lVert\tilde{\theta}_{n,2} - \theta_*\rVert\right).
\end{align*}
Finally, 
the Taylor expansion of $\sqrt{n}(\dot{\tilde{R}}_{n,1}(\tilde{\theta}_{n,2}) - \dot{\tilde{R}}_{n,2}(\tilde{\theta}_{n,1}))$ writes:
\begin{align}
    \nonumber
    & \sqrt{n}\big(\dot{\tilde{R}}_{n,1}(\tilde{\theta}_{n,2}) - \dot{\tilde{R}}_{n,2}(\tilde{\theta}_{n,1})\big) \\
    \nonumber
    & =  \sqrt{n}\big(\dot{\tilde{R}}_{n,1}(\theta_*) - \dot{\tilde{R}}_{n,2}(\theta_*) + \ddot{r}(\theta_*)(\tilde{\theta}_{n,2} - \theta_*) - \ddot{r}(\theta_*)(\tilde{\theta}_{n,1} - \theta_*) \big) + o_P(1) \\
         & =  2\sqrt{n}(\ddot{\tilde{R}}_{n,1}(\theta_*) - \ddot{\tilde{R}}_{n,2}(\theta_*)) + o_P(1) ,
\label{eq:stopcrit}
\end{align}
because $\sqrt{n} \lVert\tilde{\theta}_{n,j} - \theta_*\rVert =O_P(1)$ for $j=1,2$.
The right-hand side of equation~(\ref{eq:stopcrit}) weakly converges towards the centered Gaussian distribution with covariance matrix $8V(q_{\theta_*}, \dot{\ell}_{\theta_*})$.


\bibliography{mybibfile}

\end{document}